\documentclass[11pt]{article}

\usepackage{psfrag,amsmath,amsbsy,amsfonts,amssymb,amsthm,dsfont,fullpage,units}
\usepackage{graphicx}
\usepackage{algorithm,algorithmic,mathtools}
\usepackage{color,cases}
\usepackage{tikz}
\usetikzlibrary{calc,shapes}
\usepackage{subfig}
\usepackage{hyperref}
\usepackage{epstopdf}

\newtheorem{corollary}{Corollary}[section]

\newtheorem{lemma}{Lemma}[section]

\newtheorem{theorem}{Theorem}[section]
\theoremstyle{definition}

\newtheorem{definition}{Definition}

\newcommand{\citep}{\cite}
\newcommand{\citet}{\cite}

\newcommand\E{\mathbb{E}}
\newcommand\R{\mathbb{R}}
\newcommand\N{\mathcal{N}}

\newcommand{\Aold}{\ensuremath{\widetilde{A}}}
\newcommand{\Anew}{\ensuremath{A}}
\newcommand{\Xold}{\ensuremath{\widetilde{X}}}
\newcommand{\Xnew}{\ensuremath{X}}

\newcommand{\Astar}{\ensuremath{A^*}}
\newcommand{\Xstar}{\ensuremath{X^*}}
\newcommand{\Ystar}{\ensuremath{Y}}
\newcommand{\spindex}{\ensuremath{s}}

\newcommand{\delX}{\triangle X}
\newcommand{\row}[2]{{{#1}}^{#2}}
\newcommand{\col}[2]{{#1}_{#2}}
\newcommand{\elt}[3]{{{#1}}^{#2}_{#3}}

\newcommand{\pinv}[1]{{#1}^+}
\newcommand{\inv}[1]{{#1}^{-1}}
\newcommand{\trans}[1]{{#1}^\top}
\newcommand{\ones}{\mathrm{1}}
\newcommand{\eye}{\mathbb{I}}

\newcommand{\Aiter}[1][t]{\ensuremath{A(#1)}}
\newcommand{\Aiteri}[1][t]{\ensuremath{A_i(#1)}}

\newcommand{\Xiter}[1][t]{\ensuremath{X(#1)}}

\newcommand{\errt}[1][t]{\ensuremath{\epsilon_{#1}}}
\newcommand{\errtplus}[1][t+1]{\ensuremath{\epsilon_{#1}}}
\newcommand{\thresh}[2]{\ensuremath{\mathrm{T}_{#1}({#2})}}

\newcommand{\expec}[1]{\mathbb{E}\left[#1\right]}

\newcommand{\twonorm}[1]{\left\| {#1} \right\|_2}
\newcommand{\iprod}[2]{\langle #1, #2 \rangle}
\newcommand{\frob}[1]{\left\|#1\right\|_F}
\newcommand{\set}[1]{\left\{#1\right\}}
\newcommand{\order}[1]{\ensuremath{\mathcal{O}\left(#1\right)}}
\newcommand{\distop}[2]{\ensuremath{dist\left(#1,#2\right)}}
\newcommand{\defas}{:=}

\newcommand\inner[1]{\ensuremath{\langle #1 \rangle}}

\DeclareMathOperator{\supp}{Supp}

\def\tha{{\mbox{\tiny th}}}

\DeclarePairedDelimiter\norm{\lVert}{\rVert}

\DeclarePairedDelimiter\onenorm{\lVert}{\rVert_1}

\DeclarePairedDelimiter\infnorm{\lVert}{\rinfnorm}

 \DeclarePairedDelimiter\abs{\lvert}{\rvert}

 \def\0{{\bf 0}}

\def\viz{{viz.,\ \/}}

\def\st{{s.t.  }}

\def\qed{\hfill\hbox{${\vcenter{\vbox{
    \hrule height 0.4pt\hbox{\vrule width 0.4pt height 6pt
    \kern5pt\vrule width 0.4pt}\hrule height 0.4pt}}}$}}

\definecolor{myred}{rgb}{0.3,0.0,0.7}
\definecolor{dkg}{rgb}{0.1,0.7,0.2}
\definecolor{dkb}{rgb}{0.0,0.2,0.8}

\newcommand{\bprfof}{\begin{proof_of}}
\newcommand{\eprfof}{\end{proof_of}}
\newcommand{\bprf}{\begin{myproof}}
\newcommand{\eprf}{\end{myproof}}
\newcommand{\bp}{\begin{psfrags}}
\newcommand{\ep}{\end{psfrags}}
\newcommand{\bl}{\begin{lemma}}
\newcommand{\el}{\end{lemma}}
\newcommand{\bt}{\begin{theorem}}
\newcommand{\et}{\end{theorem}}
\newcommand{\bc}{\begin{center}}
\newcommand{\ec}{\end{center}}
\newcommand{\bi}{\begin{itemize}}
\newcommand{\ei}{\end{itemize}}
\newcommand{\ben}{\begin{enumerate}}
\newcommand{\een}{\end{enumerate}}
\newcommand{\bd}{\begin{definition}}
\newcommand{\ed}{\end{definition}}
\def\beq{\begin{equation}}
\def\eeq{\end{equation}\noindent}
\def\beqn{\begin{eqnarray}}
\def\eeqn{\end{eqnarray} \noindent}
\def\beqnn{  \begin{eqnarray*}}
\def\eeqnn{\end{eqnarray*}  \noindent}
\def\bcase{  \begin{numcases}}
\def\ecase{\end{numcases}   \noindent}
\def\bsbcase{  \begin{subnumcases}}
\def\esbcase{\end{subnumcases}   \noindent}

\newenvironment{myproof}{\noindent{\bf Proof:} \hspace*{1em}}{
    \hspace*{\fill} $\Box$ }
\newenvironment{proof_of}[1]{\noindent {\bf Proof of #1: }}{\hspace*{\fill} $\Box$ }

\newcommand{\matplottc}[1]{               
        \unitlength .45truein
        \begin{center}
        \includegraphics{#1.ps}
        \end{picture}
        \end{center}
}

\def\psfancypar#1#2{\begingroup\def\par{\endgraf\endgroup\lineskiplimit=0pt}
               \setbox2=\hbox{\large\sc #2}
               \newdimen\tmpht \tmpht \ht2 \advance\tmpht by \baselineskip
               \font\hhuge=Times-Bold at \tmpht
               \setbox1=\hbox{{\hhuge #1}}
               \count7=\tmpht \count8=\ht1
               \divide\count8 by 1000 \divide\count7 by \count8
               \tmpht=.001\tmpht\multiply\tmpht by \count7
               \font\hhuge=Times-Bold at \tmpht
               \setbox1=\hbox{{\hhuge #1}}
               \noindent
                \hangindent1.05\wd1
               \hangafter=-2 {\hskip-\hangindent
               \lower1\ht1\hbox{\raise1.0\ht2\copy1}%
                \kern-0\wd1}\copy2\lineskiplimit=-1000pt}

\def\Kout{\setbox1=\hbox{\Huge\bf K}\hbox to
1.05\wd1{\hspace{.05\wd1}
}}
\def\Sout{\setbox1=\hbox{\Huge\bf S}\hbox to 1.05\wd1{\hspace{.05\wd1}
}}

\renewcommand{\abs}[1]{\left|#1\right|}
\renewcommand{\infnorm}[1]{\left\|#1\right\|_{\infty}}

\newtheorem*{rep@theorem}{\rep@title}
\newcommand{\newreptheorem}[2]{%
\newenvironment{rep#1}[1]{%
 \def\rep@title{#2 \ref{##1}}%
 \begin{rep@theorem}}%
 {\end{rep@theorem}}}
\newreptheorem{theorem}{Theorem}

\newtheorem*{rep@lemma}{\rep@title}
\newcommand{\newreplemma}[2]{%
\newenvironment{rep#1}[1]{%
 \def\rep@title{#2 \ref{##1}}%
 \begin{rep@lemma}}%
 {\end{rep@lemma}}}
\newreptheorem{lemma}{Lemma}

\newtheorem*{rep@prop}{\rep@title}
\newcommand{\newrepprop}[2]{%
\newenvironment{rep#1}[1]{%
 \def\rep@title{#2 \ref{##1}}%
 \begin{rep@prop}}%
 {\end{rep@prop}}}
\newreptheorem{prop}{Proposition}

\DeclareMathOperator{\nzset}{Supp}
\newcommand{\term}{\ensuremath{\mathcal{T}}}
\newcommand{\singmin}{\ensuremath{\sigma_{\min}}}
\newcommand{\singmax}{\ensuremath{\sigma_{\max}}}
\renewcommand{\P}{\ensuremath{\mathbb{P}}}

\newcommand{\Xoip}{\elt{\Xstar}{p}{i}}
\newcommand{\spmat}{\ensuremath{\chi}}
\newcommand{\dip}{\elt{\spmat}{p}{i}}
\newcommand{\Mip}{\elt{M}{p}{i}}
\newcommand{\diq}{\elt{\spmat}{q}{i}}
\newcommand{\Miq}{\elt{M}{q}{i}}

\renewcommand{\P}{\ensuremath{\mathbb{P}}}
\title{Learning Sparsely Used Overcomplete Dictionaries \\via
  Alternating Minimization}

\author{Alekh Agarwal, Animashree Anandkumar, Prateek Jain,
  \\ Praneeth Netrapalli\,\footnote{A. Agarwal is with Microsoft
    Research, New York, USA. Email:
    alekha@microsoft.com. A. Anandkumar is with the Center for
    Pervasive Communications and Computing, Electrical Engineering and
    Computer Science Dept., University of California, Irvine, USA
    92697. Email: a.anandkumar@uci.edu. P. Jain is with Microsoft
    Research, Bangalore, India. Email:
    prajain@microsoft.com. P. Netrapalli is with Dept. of ECE, The
    University of Texas at Austin. Email: praneethn@utexas.edu. Part
    of this work was done when A. Anandkumar and P. Netrapalli were
    visiting Microsoft Research. An extended abstract containing an
    earlier version of these results appears in COLT 2014.}}

\begin{document}
\maketitle

\begin{abstract}
We consider the problem of sparse coding, where each sample consists
of a sparse linear combination of a set of dictionary atoms, and the task is
to learn both the dictionary elements and the mixing
coefficients. Alternating minimization is a popular heuristic for
sparse coding, where the dictionary and the coefficients are estimated
in alternate steps, keeping the other fixed. Typically, the
coefficients are estimated via $\ell_1$ minimization, keeping the
dictionary fixed, and the dictionary is estimated through least
squares, keeping the coefficients fixed.  In this paper, we establish
local linear convergence for this variant of alternating minimization
and establish that the basin of attraction for the global optimum
(corresponding to the true dictionary and the coefficients) is
$\order{1/s^2}$, where $s$ is the sparsity level in each sample and
the dictionary satisfies RIP. Combined with the recent
results of approximate dictionary estimation, this yields provable
guarantees for exact recovery of both the dictionary elements and the
coefficients, when the dictionary elements are incoherent.
\end{abstract}


\section{Introduction}
A sparse code encodes each sample with a sparse set of elements,
termed as dictionary atoms. Specifically, given a set of samples $Y\in
\R^{d\times n}$, the generative model is \[ Y= \Astar \Xstar, \qquad
\Astar \in \R^{d\times r}, \Xstar\in \R^{r \times n},\] and
additionally, each column of $\Xstar$ has at most $s$ non-zero
entries. The columns of $\Astar$ correspond to the dictionary atoms,
and the columns of $\Xstar$ correspond to the mixing coefficients of
each sample. Each sample is a combination of at most $s$ dictionary
atoms.  Sparse codes can thus succinctly represent high dimensional
observed data.

The problem of sparse coding consists of unsupervised learning of the
dictionary and the coefficient matrices. Thus, given only unlabeled
data, we aim to learn the set of dictionary atoms or basis functions
that provide a good fit to the observed data. Sparse coding is applied
in a variety of domains. Sparse coding of natural images has yielded
dictionary atoms which resemble the receptive fields of neurons in the
visual cortex~\cite{olshausen1996emergence,olshausen1997sparse}, and
has also yielded localized dictionary elements on speech and video
data~\cite{lewicki2000learning,olshausen2000sparse}.

An important strength of sparse coding is that it can incorporate
overcomplete dictionaries, where the number of dictionary atoms $r$
can exceed the observed dimensionality $d$. It has been argued that
having overcomplete representation provides greater flexibility is
modeling and more robustness to noise~\cite{lewicki2000learning},
which is crucial for encoding complex signals present in images,
speech and video. It has been shown that the performance of most
machine learning methods employed downstream is critically dependent
on the choice of data representations, and overcomplete
representations are the key to obtaining state-of-art prediction
results~\cite{bengio2012unsupervised}.

On the downside, the problem of learning sparse codes is
computationally challenging, and is in general,
NP-hard~\cite{rey1994adaptive}. In practice, heuristics are employed
based on alternating minimization.  At a high level, this consists of
alternating steps, where the dictionary is kept fixed and the
coefficients are updated and vice versa. Such alternating minimization
methods have enjoyed empirical success in a number of
settings~\cite{lee2006efficient,engan1999method,aharon2006img,mairal2008discriminative,yang2010image}.
In this paper, we carry out a theoretical analysis of the alternating
minimization procedure for sparse coding.

\subsection{Summary of Results}

We consider the alternating minimization procedure where we employ an
initial estimate of the dictionary and then use $\ell_1$ based
minimization for estimating the coefficient matrix, given the
dictionary estimate. The dictionary is subsequently re-estimated given
the coefficient estimates.  We establish local convergence to the true
dictionary $\Astar$ and coefficient matrix $\Xstar$ for this procedure
whenever $\Astar$ satisfies RIP for $2s$-sparse vectors.
In other words, we characterize the ``basin of attraction'' for the true solution
$(\Astar, \Xstar)$ and establish that alternating minimization
succeeds in its recovery when a dictionary is initialized with an
error of at most $\order{1/s^2}$, where $s$ is the sparsity
level. More precisely, the initial dictionary estimate $\Aiter[0]$ is
required to satisfy
\[ \errt[0]:=\max_{i\in [r]} \min_{z\in \{-1,+1\}}
\twonorm{z\Astar_i - \Aiter[0]_i}= \order{\frac{1}{s^2}},\] where
$\Astar_i$ represents $i^{\tha}$ column of $\Astar$.

Further when the sparsity level satisfies $s = \order{d^{1/6}}$ and
the number of samples satisfies $n = \order{r^2}$, we establish a
linear rate of convergence for the alternating minimization procedure
to the true dictionary even when the dictionary is overcomplete
$(r\geq d)$, .

For the case of incoherent dictionaries,
by combining the above result with recent results on approximate
dictionary estimation by Agarwal
et. al~\cite{AgarwalAN13} or Arora
et. al~\cite{AroraICML}, we guarantee exact recovery of the true
solution $(\Astar, \Xstar)$ when the alternating procedure is
initialized with the output of~\cite{AgarwalAN13}
or~\cite{AroraICML}.  If we employ the procedure of Agarwal
et. al~\cite{AgarwalAN13}, the overall requirements are
as follows: the sparsity level is required to be $s = \order{d^{1/9},
  r^{1/8}}$, and the number of samples $n = \order{r^2}$ to
guarantee exact recovery of the true solution. If we employ the
procedure of Arora et. al~\cite{AroraICML} (in particular their
\textsc{OverlappingAverage} procedure), we can establish exact
recovery assuming $s  = \order{r^{1/6}, \sqrt{d}}$.

\subsection{Related Work}

\paragraph{Analysis of local optima of non-convex programs for sparse coding: }
Gribonval and Schnass~\cite{gribonval2010dictionary}, Geng et
al.~\cite{GengWaWr2011} and Jenatton et al.~\cite{jenatton2012local}
carry out a theoretical analysis and study the conditions under which
the true solution turns out to be a local optimum of a non-convex
optimization problem for dictionary recovery.  Gribonval and
Schnass~\cite{gribonval2010dictionary} and Geng
et. al~\cite{GengWaWr2011} both consider the noiseless setting, and
analyze the following non-convex program

\beq \label{eqn:nonconvex}\min \|X\|_1\qquad \st, \,\, Y=AX,\,\,
\|A_i\|_2=1, \,\,\forall\,i\in [r].\eeq
Since $A$ and $X$ are both unknown, the constraint $Y = AX$ is
non-convex. It is natural to expect the true solution $(\Astar,
\Xstar)$ to be a local optimum for \eqref{eqn:nonconvex} under fairly
mild conditions, but this turns out to be non-trivial to
establish. The difficulties arise from the non-convexity of the
problem and the presence of sign-permutation ambiguity which leads to
exponentially many equivalent solutions obtained via sign change and
permutation.  Gribonval and Schnass~\cite{gribonval2010dictionary} established that $(\Astar,
\Xstar)$ is a local optimum for \eqref{eqn:nonconvex}, but limited to
the case where the dictionary matrix $A$ is square and hence, did not
incorporate the overcomplete setting.
Geng et al.~\cite{GengWaWr2011} extend the analysis to the
overcomplete setting, and establish that the true solution is a local
optimum of \eqref{eqn:nonconvex} w.h.p. for incoherent dictionaries,
when the number of samples $n$ and sparsity level $s$ scale as

\beq \label{eqn:gengresult}n = \Omega\left(\|A\|_2^4 r^3 s \right),\quad
s =\order{\sqrt{d}}.\eeq
In our setting, where the spectral norm is assumed to be $\|A\|_2 <
\mu_1 \sqrt{r/d}$, for some constant $\mu_1>0$, the sample complexity
simplifies as $n = \Omega\left( r^5 s/d^2 \right)$. Jenatton et
al.~\cite{jenatton2012local} consider the noisy setting and analyze
the modified non-convex program involving $\ell_1$ penalty for the
coefficient matrix and $\ell_2$ penalty for the loss in fitting the
samples, and establish that the true solution is in the neighborhood
of a local optimum of the modified non-convex program w.h.p. when the
number of samples scales as $n =\Omega\left(\|A\|_2^2 r^3 d s^2
\right)$. In our setting, this reduces to $n = \Omega \left( r^4
s^2\right) $.  There are significant differences of the above works
from ours. While these works establish that $(\Astar, \Xstar)$ is a
local optimum of a non-convex program, they do not provide a tractable
algorithm to reach this particular solution as opposed to another
local optimum. In contrast, we establish guarantees for a simple
alternating minimization algorithm and explicitly characterize the
``basin of attraction'' for the true solution $(\Astar, \Xstar)$. This
provides precise initialization conditions for the alternating
minimization to succeed. Moreover, our sample complexity requirements
are much weaker and we require only $n = \order{r^2}$ samples for
our guarantees to hold.

\paragraph{Alternating minimization for sparse coding: }
Our analysis in this paper provides a theoretical explanation for the
empirical success of alternating minimization, observed in a number of
works~\cite{lee2006efficient,engan1999method,aharon2006img,mairal2008discriminative,yang2010image}. These
methods are all based on alternating minimization, but differ mostly
in how they update the dictionary elements. For instance, Lee
et. al. carry out least squares for updating the
dictionary~\cite{lee2006efficient} similar to the the method of
optimal directions~\cite{engan1999method}, while the K-SVD
procedure~\cite{aharon2006img}, updates the dictionary estimate using
a spectral procedure on the residual. However, none of the previous
works provide theoretical guarantees on the success of the alternating
minimization procedure for sparse coding.

\paragraph{Guaranteed dictionary estimation: }
Some of the recent works provide theoretical guarantees on the
estimation of the true dictionary. Spielman
et. al~\cite{spielman2012exact} establish exact recovery under
$\ell_1$ based optimization when the true dictionary $\Astar$ is a
basis, which rules out the overcomplete setting.  Agarwal
et. al~\cite{AgarwalAN13} and Arora
et. al~\cite{AroraICML} propose methods for approximate dictionary
estimation in the overcomplete setting. At a high level, both their
methods involve a clustering-based approach for finding samples which
share a dictionary element, and then using the subset of samples to
estimate a dictionary element.   Agarwal
et. al~\cite{AgarwalAN13} establish exact recovery of
the true solution $(\Astar, \Xstar)$ under a ``one-shot'' Lasso
procedure, when the non-zero coefficients are Bernoulli $\{-1,+1\}$ (or
more generally discrete). On the other hand, we assume only   mild conditions on the
non-zero elements. Arora
et. al~\cite{AroraICML} consider an alternating minimization procedure. However, a key distinction is that their analysis requires {\em fresh} samples in each iteration, while we consider the same samples for all the iterations. We show {\em exact} recovery using $n =\Omega(r^2)$ samples, while~\cite{AroraICML} can only establish that the error is bounded by $\exp[-O(n/r^2)]$.
Furthermore, both the above papers~\cite{AroraICML,AgarwalAN13}
assume that the dictionary elements are mutually incoherent. Our local
convergence result in this paper assumes only that the dictionary matrix satisfies
RIP (which is strictly weaker than incoherence). For the case of incoherent
dictionaries, we can employ the procedures of
Agarwal et. al~\cite{AgarwalAN13} or Arora
et. al~\cite{AroraICML} for initializing the alternating procedure and
obtain overall guarantees in such scenarios.

\paragraph{Other works on sparse coding: }
Some of the other recent works are only tangentially related to this
paper. For instance, the works~\cite{vainsencher2011sample,
  mehta2013sparsity,maurer2012sparse,TRS:dict2013} provide
generalization bounds for predictive sparse coding, without
computational considerations, which differs from our generative
setting here and algorithmic considerations.  Parametric dictionary
learning is considered in~\cite{yaghoobi2009parametric}, where the
data is fitted to dictionaries with small coherence. Note that we
provide guarantees when the underlying dictionary is incoherent, but
do not constrain our method to produce an incoherent dictionary. The
problem of sparse coding is also closely related to the problem of
blind source separation, and we refer the reader
to~\cite{AgarwalAN13} for an extended survey of these
works.

\paragraph{Majorization-minimization algorithms for biconvex optimization:}
Beyond the specific problem of sparse coding, alternating optimization
procedures more generally are a natural fit for biconvex optimization
problems, where the objective is individually convex in two sets of
variables but not jointly convex. Perhaps the most general study of
these problems has been carried out in the framework of
majorization-minimization schemes~\cite{LangeHuYa2000}, or under the
name of the EM algorithm in statistics literature. In this generality,
the strongest result one can typically provide is a convergence
guarantee to a local optimum of the problem. When the bi-convex
objective is defined over probability measures, Csiszar presents a
fairly general set of conditions on the objective function, under
which linear convergence to the global optimum is guaranteed (see,
e.g. the recent tutorial~\cite{CsiszarSh2004} for an excellent
overview). However, these conditions do not seem to easily hold in the
context of dictionary learning. Alternating optimization in related
contexts has also been studied in a variety of matrix factorization
problems such as low-rank matrix completion and non-negative matrix
factorization. Perhaps the most related to our work are similar
results for low-rank matrix completion problems by Jain et
al.~\cite{JainNeSa2013}. 


\paragraph{Notation: }
Let $[n]:=\{1,2, \ldots, n\}$. For a vector $v$ or a matrix $W$, we
will use the shorthand $\nzset(v)$ and $\nzset(W)$ to denote the set
of non-zero entries of $v$ and $W$ respectively.  $\|w\|_p$ denote
the $\ell_p$ norm of vector $w$; by default, $\|w\|$ denotes $\ell_2$ norm of $w$. 
$\|W\|_2$ denotes the spectral norm (largest singular value) of matrix $W$. $\|W\|_\infty$ denotes the largest element (in magnitude) of $W$. For a matrix $X$,
$\row{X}{i}$, $\col{X}{i}$ and $\elt{X}{i}{j}$ denote the
$i^{\tha}$ row, $i^{\tha}$ column and $(i,j)^{\tha}$ element of $X$
respectively.

\section{Algorithm}
\label{sec:method}

Given an initial estimate of the dictionary, we alternate
between two procedures, \viz a sparse recovery step for estimating the
coefficients given a dictionary, and a least squares step for a
dictionary given the estimates of the coefficients. The details of
this approach are presented in Algorithm~\ref{algo:altmin}.
\floatname{algorithm}{Algorithm}
\begin{algorithm}[t]
  \caption{AltMinDict$(Y, \Aiter[0], \errt[0])$: Alternating
    minimization for dictionary learning}
  \label{algo:altmin}
  \begin{algorithmic}[1]
    \renewcommand{\algorithmicrequire}{\textbf{Input: }}
    \renewcommand{\algorithmicensure}{\textbf{Output: }}
    \REQUIRE Samples $Y$, initial dictionary estimate $\Aiter[0]$, accuracy
    sequence $\errt$ and sparsity level $s$. Thresholding function
    $\thresh{\rho}{a}= a$ if $|a|>\rho$ and $0$ o.w.
    \FOR{iterations $t = 0,1,2,\ldots, T-1$}
    \FOR{samples $i = 1,2,\ldots, n$}
    \STATE $\col{\Xiter[t+1]}{i} = \arg\min_{x \in \R^r} \norm{x}_1$\quad
    such that, $\norm{Y_i - \Aiter x}_2 \leq
    \errt$
    \ENDFOR
    \STATE Threshold: $\Xiter[t+1] = \Xiter[t+1]\cdot *(\mathbb{I}[\Xiter[t+1]>9s\errt])$
    \STATE Estimate $\Aiter[t+1] =Y\pinv{\Xiter[t+1]}$
    \STATE Normalize: $\col{\Aiter[t+1]}{i} =
    \frac{\col{\Aiter[t+1]}{i}}{\norm{\col{\Aiter[t+1]}{i}}_2} $
    \ENDFOR
    \ENSURE $\Aiter[T]$
  \end{algorithmic}
\end{algorithm}


The sparse recovery step of Algorithm~\ref{algo:altmin} is based on
$\ell_1$-regularization, followed by thresholding.  The thresholding
is required for us to guarantee that the support set of our
coefficient estimate $\Xiter$ is a {\em subset} of the true support
with high probability. Once we have an estimate of the coefficients,
the dictionary is re-estimated through least squares.  The overall
algorithmic scheme is popular for dictionary learning, and there are a
number of variants of the basic method. For instance, the
$\ell_1$-regularized problem in step 3 can also be replaced by other
robust sparse recovery procedures such as OMP~\cite{Tropp2007} or
GraDeS~\cite{GargK09}. More generally the exact lasso and
least-squares steps may be replaced with other optimization methods
for computational efficiency, e.g.~\cite{jenatton2010proximal}.



\section{Main results and their proofs}\label{sec:guarantees}
In this section, we provide our local convergence result for alternating
minimization and also clearly
specify all the required assumptions on $\Astar$ and $\Xstar$.
We provide a brief
sketch of our proof for each of the steps in
Section~\ref{sec:overview}.

\subsection{Assumptions}

We start by formally describing the assumptions needed for the main
recovery result of this paper. Without loss of
  generality, assume that all the elements are normalized:
  $\|\Astar_i\|_2 =1$, for $i \in [r]$. This is because we can always rescale the dictionary elements
and the corresponding coefficients and obtain the same observations.

\paragraph{Assumptions: }

\ben
\item[$(A1)$] {\bf Dictionary Matrix satisfying RIP: }
 The dictionary matrix $\Astar$ has a $2s$-RIP constant of $\delta_{2s}<0.1$.

\item[$(A2)$] {\bf Spectral Condition on Dictionary Elements: }The
  dictionary matrix has bounded spectral norm, for some constant $\mu_1>0$,
  $\|\Astar \|_2 < \mu_1\sqrt{\frac{r}{d}}$.

\item[$(A3)$] {\bf Non-zero Entries in Coefficient Matrix: }We assume
  that the non-zero entries of $\Xstar$ are drawn i.i.d. from a
  distribution such that
  $\expec{\left(\elt{\Xstar}{i}{j}\right)^2}=1$, and satisfy the
  following a.s.: $|\elt{\Xstar}{i}{j}|\leq M, \forall i,j$.

\item[$(A4)$] {\bf Sparse Coefficient Matrix: }The columns of
  coefficient matrix have $s$ non-zero entries which are selected
  uniformly at random from the set of all $s$-sized subsets of $[r]$,
  i.e. $|\supp(\Xstar_i)| =s$, $\forall\, i \in [n]$.
  We require $s$ to satisfy $s< \frac{d^{1/6}}{c_2 \mu_1^{1/3} }$, for some universal constant $c_2$.



\item[$(A5)$] {\bf Sample Complexity: } For some universal constant
  $c> 0$ and a given failure parameter $\delta>0$, the number of
  samples $n$ needs to satisfy
  \begin{align*}
    &n \geq c_3\,\max\left(  r^2 , rM^2s\right)
    \log \frac{2r}{\delta}, \quad ,
  \end{align*} where    $c_3>0$ is a universal constant.

\item[$(A6)$]{\bf Initial dictionary with guaranteed error bound: } We
  assume that we have access to an initial dictionary estimate
  $\Aiter[0]$ such that\[ \widehat{\errt[0]}:=\max_{i\in [r]} \min_{z\in
    \{-1,+1\}} \twonorm{z\Aiteri[0] - \Astar_i}< \frac{1}{2592s^2}.\]

\item[$(A7)$] {\bf Choice of Parameters for Alternating Minimization:}
  Algorithm~\ref{algo:altmin} uses a sequence of accuracy parameters
  $\errt[0]=1/2592 s^2 $ and

  \begin{equation} \errt[t+1] = \frac{25050 \mu_1
      s^3}{\sqrt{d}} \errt.
    \label{eqn:errt-seq}
  \end{equation}

\een

Assumption $(A1)$ regarding the RIP assumption is crucial in establishing our guarantees, since it
is critical for analyzing the performance of the compressed sensing
subroutine in Algorithm~\ref{algo:altmin} (steps 2-5). It is possible
to further weaken this assumption to a Restricted Eigenvalue condition
which is often used in the sparse regression literature as well~\cite{RaskuttiWaYu2010,NegRavWaiYu09}. We
will present a more detailed discussion of this condition in the proof
sketch. In order to keep the results with cleaner constants, we will
continue with the RIP assumption for the rest of the analysis, while
mentioning how the result can be extended easily under a more general
restricted eigenvalue assumption.

The assumption $(A2)$ provides a bound on the spectral norm of
$\Astar$. Note that the RIP and spectral assumptions are satisfied
with high probability (w.h.p.)  when the dictionary elements are
randomly drawn from a mean-zero sub-gaussian distribution.

Assumption $(A3)$ imposes some natural constraints on the non-zero
entries of $\Xstar$. Assumption$(A4)$ on sparsity in the coefficient
matrix is crucial for identifiability of the dictionary learning
problem.


Assumption $(A5)$ provides a bound on sample complexity.
Assumption $(A6)$ specifies the accuracy of the initial estimate required
by Algorithm~\ref{algo:altmin}. Recent works \cite{Arora2013,AgarwalAN13}
provide provable ways of obtaining such an estimate. Please see Section~
\ref{sec:combine-guarantees} for more details.

Assumption $(A7)$ specifies the choice of accuracy parameters used by
alternating method in Algorithm~\ref{algo:altmin}. Due to Assumption
$(A4)$ on sparsity level $s$, we have that $\frac{25050 \mu_1
  s^3}{\sqrt{d}} <1/2$ and the accuracy parameters in
\eqref{eqn:errt-seq} form a decreasing sequence. This implies that in
Algorithm~\ref{algo:altmin}, the accuracy constraint becomes more
stringent with the iterations of the alternating method.

\subsection{Guarantees for Alternating Minimization}
\label{sec:guarantee_altmin}
We now prove a local convergence result for alternating
minimization. We assume that we have access to a good initial estimate
of the dictionary:

\begin{theorem}[Local linear convergence]\label{thm:main-altmin}
  Under assumptions $(A1)$-$(A7)$, with probability at least \mbox{$1-
    2\delta$} the iterate $\Aiter$ of Algorithm~\ref{algo:altmin}
  satisfies the following for all $t \geq 1$:
\begin{align*}
  \min_{z \in \{-1,1\}} \norm{z\Aiteri    - \Astar_i}_2 \leq \sqrt{2}\errt, 1\leq i\leq r.
\end{align*}
\end{theorem}

\paragraph{Remarks: }
Note that we have a sign ambiguity in recovery of the dictionary
elements, since we can exchange the signs of the dictionary elements
and the coefficients to obtain the same observations.

Theorem~\ref{thm:main-altmin} guarantees that we can recover the
dictionary $\Astar$ to an arbitrary precision $\epsilon$ (based on the
number of iterations $T$ of Algorithm~\ref{algo:altmin} ), given $n =
\order{r^2}$ samples. We contrast this with the results
of~\citet{Arora2013}, who also provide recovery guarantees to an
arbitrary accuracy $\epsilon$, but only if the number of samples is
allowed to increase as $\order{r^2 \log 1/\epsilon}$.

The consequences of Theorem~\ref{thm:main-altmin} are powerful
combined with our Assumption $(A4)$ and the
recurrence~\ref{eqn:errt-seq} (since $(A4)$ ensures that $\errt$ forms
a decreasing sequence). In particular, it is implied that with high
probability we obtain,
\begin{equation*}
  \min_{z \in \{-1,1\}}\norm{z\Aiteri[t] - \col{\Astar}{i}}_2 \leq
  \widehat{\errt[0]} 2^{-t}.
\end{equation*}
Given the above bound, we need at most $\order{
  \log_2\frac{\widehat{\errt[0]}}{\epsilon}}$ in order to ensure
$\norm{z\Aiteri[T] - \col{\Astar}{i}}_2 \leq \epsilon$ for all
the dictionary elements $i = 1,2,\ldots, r$. In the convex
optimization parlance, the result demonstrates a local linear
convergence of Algorithm~\ref{algo:altmin} to the globally optimal
solution under an initialization condition. Another way of
interpreting our result is that the global optimum has a \emph{basin
  of attraction} of size $\order{1/s^2}$ for our alternating
minimization procedure under these assumptions (since we require
$\widehat{\errt[0]} \leq \order{1/s^2}$).

We also recall that the lasso step in
Algorithm~\ref{algo:altmin} can be replaced with a different robust
sparse recovery procedure, with qualitatively similar results.

\subsection{Using Local Convergence for Complete Recovery}
\label{sec:combine-guarantees}
In the above section, we showed a local convergence result for
Algorithm~\ref{algo:altmin}.  In particular, Assumption~$(A6)$
requires that the initial dictionary estimate be at most
$\order{\frac{1}{s^2}}$ away from $\Astar$.  In this section, we use
the recent result of \cite{AgarwalAN13} to obtain an initialization which satisfies  Assumption
$(A6)$, and thus, we obtain a full recovery result for the sparsely-used
dictionary problem with assumptions only on the model parameters.  In
order to obtain the initialization from the method of~\cite{AgarwalAN13}, we require the
following assumptions:

\begin{itemize}

\item[$(B1)$] {\bf Incoherent Dictionary Elements: }Without loss of
  generality, assume that all the elements are normalized:
  $\|\Astar_i\|_2 =1$, for $i \in [r]$.
  We assume pairwise incoherence
  condition on the dictionary elements, for some constant $\mu_0>0$,
  $|\inner{\Astar_i, \Astar_j}| < \frac{\mu_0}{\sqrt{d}}$.

\item[$(B3)$] {\bf Non-zero Entries in Coefficient Matrix: }We assume
  that the non-zero entries of $\Xstar$ are drawn i.i.d. from a
  distribution such that
  $\expec{\left(\elt{\Xstar}{i}{j}\right)^2}=1$, and satisfy the
  following a.s.: $m \leq |\elt{\Xstar}{i}{j}|\leq M, \forall i,j$.

\item[$(B4)$] {\bf Sparse Coefficient Matrix: }The columns of
  coefficient matrix have bounded number of non-zero entries $s$ which
  are selected randomly, i.e.  \beq \label{eqn:sparsity} |\supp(x_i)|
  =s, \quad \forall \, i \in [n].\eeq

    We require $s$ to be
  \[
    s < c_1 \min\left(\frac{m}{M}\frac{d^{1/4}}{\sqrt{\mu_0}}, \left(\frac{d}{\mu_1^2} \frac{m^4}{M^4}\right)^{1/9},
	r^{1/8}\left(\frac{m}{M}\right)^{1/4}\right),
  \]for universal constants $c_1, c_2>0$. Constants $m, M$ are as specified above.

\item[$(B5)$] {\bf Sample Complexity: }Given universal constant $c_2>0$,
  choose $\delta>0$ and the number of samples $n$ such that

  \[
  n := n(d, r, s, \delta) \geq c_2\,r^2\frac{M^2}{m^2} \log \frac{2r}{\delta}.
  \]
\end{itemize}

\begin{theorem}[Specialization of Theorem~2.1 from \cite{AgarwalAN13}]
  Under assumptions $(B1),\, (A2),\, (B3)-(B5)$ and $(A7)$, there exists
  an algorithm which given $Y$ outputs $\Aiter[0]$, such that
  Assumption $(A6)$ holds with probability greater than
  $1-2n^2\delta$.
\end{theorem}

The restatement follows by setting $\alpha = s^{-9/2} \frac{m^2}{M^2}$ in that result
which ensures that the error in the initialization is at most $1/s^2$
as required by Assumption $(A6)$. Combining the above theorem with
Theorem~\ref{thm:main-altmin} gives the following powerful corollary.

\begin{corollary}[Exact recovery]
\label{coro:full-recovery}
Suppose assumptions $(B1),\, (A2)-(A5),\, (B3)-(B5)$ and $(A7)$ hold. If we
start Algorithm~\ref{algo:altmin} with the output of Algorithm~1 of
\cite{AgarwalAN13}, then the following holds for all $t \geq 1$:
\begin{align*}
  \min_{z \in \{-1,1\}} \norm{z\Aiteri    - \Astar_i}_2 \leq \sqrt{2} \errt, 1\leq i\leq r.
\end{align*}
\end{corollary}

The above result makes use of Lemma~\ref{lem:RIP-A} in the appendix
which shows that Assumptions $(B1)$ and $(B4)$ imply $(A1)$.  Note
that the above corollary gives an exact recovery result with the only
assumptions being those on the model parameters. We also note that the
conclusion of Corollary~\ref{coro:full-recovery} does not crucially
rely on initialization specifically by the output of Algorithm~1 of
\cite{AgarwalAN13}, and admits any other initialization satisfying
Assumption $(A6)$.  As remarked earlier, the recent work
of~\citet{Arora2013} provides an alternative initialization strategy
for our alternating minimization procedure. Indeed, under our sample
complexity assumption, their \textsc{OverlappingAverage} method
provides a solution with $\widehat{\errt[0]} = \order{s/\sqrt{r}}$
assuming $s = \order{\max(r^{2/5}, \sqrt{d})}$. In particular, if $s =
\order{r^{1/6}}$, we obtain the desired initial error of $1/s^2$ using
that algorithm. The sample complexity of the entire procedure remains
identical to that in Assumption $(A5)$.


\subsection{Overview of Proof}
\label{sec:overview}
In this section we outline the key steps in proving
Theorem~\ref{thm:main-altmin}.

For ease of notation, let us consider just one iteration of
Algorithm~\ref{algo:altmin} and denote $\Xiter[t+1]$ as $\Xnew$,
$\Aiter[t+1]$ as $\Anew$ and $\Aiter[t]$ as $\Aold$. Then we have the
least-squares update:
\begin{align*}
  \Anew - \Astar &= Y\pinv{\Xnew} - \Astar\\ &= \Astar \Xstar
  \pinv{\Xnew} - \Astar \Xnew \pinv{\Xnew} = \Astar\delX\pinv{\Xnew},
\end{align*}
where $\delX = \Xstar - \Xnew$. This means that we can understand the
error in dictionary recovery by the error in the least squares
operator $\delX\pinv{\Xnew}$. In particular, we can further expand the
error in a column $p$ as: 
\begin{align*}
  \col{\Anew}{p} - \col{\Astar}{p} &=
  \col{\Astar}{p}\elt{(\delX\pinv{\Xnew})}{p}{p} +
  \col{\Astar}{\setminus p} \elt{(\delX\pinv{\Xnew})}{\setminus p}{p},
\end{align*}
where the notation $\setminus p$ represents the collection of all
indices apart from $p$, i.e., $\col{\Astar}{\setminus p}$ denote all the columns of $A$ except the $p$-th column and $\elt{(\delX\pinv{\Xnew})}{\setminus p}{p}$ denotes the off-diagonal elements of the $p$-th column of $(\delX\pinv{\Xnew})$. 

The above equation indicates that there are two sources of error in our
dictionary estimate. The element $\elt{(\delX\pinv{\Xnew})}{p}{p}$
causes the rescaling of $\col{\Anew}{p}$ relative to
$\col{\Astar}{p}$. However, this is a minor issue since the
renormalization would correct it.

More serious is the contribution from the off-diagonal terms
$\elt{(\delX\pinv{\Xnew})}{p}{\setminus p}$, which corrupt our
estimate $\col{\Anew}{p}$ with other dictionary elements beyond
$\col{\Astar}{p}$. Indeed, a crucial argument in our proof is
controlling the contribution of these terms at an appropriately small
level. In order to do that, we start by controlling the magnitude of
$\delX$.

\begin{lemma}[Error in sparse recovery]
\label{lem:update-X}
  Let $\delX \defas \Xiter[t] - \Xstar$. Assume that $2\mu_0s/\sqrt{d}
  \leq 0.1$ and $\sqrt{s\errt} \leq 0.1$ Then, we have $\nzset(\delX)
  \subseteq \nzset(\Xstar)$ and the error bound $\infnorm{\delX} \leq
  9s\errt$.
\end{lemma}

\paragraph{More general Restricted Eigenvalue conditions:}
The above  lemma is the only part of  our proof, where we   require the RIP assumption. This is  in
order to invoke the result of Candes~\cite{Candes2008} regarding the
error in compressed sensing with (bounded) deterministic noise. Such
results can also be typically established under weaker Restricted
Eigenvalue assumptions (henceforth RE). Given a vector $v \in \R^r$,
these RE conditions study the norms $\|Av\|_2^2$. A particular form of
RE condition for these approximately sparse vectors then posits (see
e.g.~\cite{RaskuttiWaYu2010,NegRavWaiYu09})

\begin{equation}
  \norm{Av}_2 \geq \gamma \norm{v}_2 - \tau \norm{v}_1.
  \label{eqn:RE}
\end{equation}
Under such a condition, it can be readily shown that
Lemma~\ref{lem:update-X} continues to hold with an error bound which
is $\order{s\errt/(\gamma - s\tau)}$. For many random matrix ensembles
for $A$, it is known that $\tau = \order{\sqrt{(\log r)/d}}$, which
means that $\gamma - s\tau$ will be bounded away from zero under our
assumptions on the sparsity level $s$. These RE conditions are the
weakest known conditions under which compressed sensing using
efficient procedures is possible, and we employ this as a subroutine
in our alternating minimization procedure.

The next lemma is very useful in our error analysis, since we
establish that any matrix $W$ satisfying $\nzset(W) \subseteq
\nzset(\Xstar)$ has a good bound on its spectral norm (even if the
entries depend on $\Astar, \Xstar$).

\begin{lemma}\label{lem:W-spectralnorm}
With probability at least $1-r\exp\left(-\frac{C n}{rs}\right)$, for
every $r \times n$ matrix $W$ s.t. $\supp(W)\subseteq \supp(\Xstar)$,
we have
\begin{align*}
  \twonorm{W} \leq 2\|W\|_\infty\sqrt{\frac{s^2n}{r}}.
\end{align*}
\end{lemma}

A particular consequence of this lemma is that it guarantees the
invertibility of the matrix $\Xnew\trans{\Xnew}$, so that the
pseudo-inverse $\pinv{\Xnew}$ is well-defined for subsequent least
squares updates. Next, we present the most crucial step which is
controlling the off-diagonal terms
$\elt{(\delX\pinv{\Xnew})}{p}{\setminus p}$.

\begin{lemma}[Off-diagonal error bound]
  \label{lem:bound-XXstarplus}
  With probability at least $1 -
  r\exp\left(-\frac{C n}{rM^2s}\right) -r \exp\left(-Cn/r^2\right)$,
  we have uniformly for every $p \in [r]$ and every $\delX$ such that
  $\infnorm{\delX} < \frac{1}{288 \spindex}$.
\begin{align*}
  \twonorm{\elt{\left(\delX \pinv{\Xnew}\right)}{\setminus p}{p}} =
  \twonorm{\elt{\left(\Xstar \pinv{\Xnew}\right)}{\setminus p}{p}}
  \leq \frac{1968 \spindex^2 \infnorm{\delX} }{\sqrt{r}}.
\end{align*}
\end{lemma}

The lemma uses the earlier two lemmas along with a few other auxilliary
results. Given these lemmas, the proof of the main theorem follows
using basic linear algebra arguments. Specifically, for any unit vector $w$ such that $w
\perp \col{\Astar}{p}$, we can bound the normalized inner product
$\inner{w, \col{\Anew}{p}}/\norm{\col{\Anew}{p}}_2$ which suffices to
obtain the result of the theorem. 

\subsection{Detailed Proof of Theorem~\ref{thm:main-altmin}}
We now provide a proof of the theorem using the above given lemmas. The proofs
of the lemmas are deferred to the appendix. Recall that, we denote
$\Aiter[t]$ as $\Aold$ and $\Aiter[t+1]$ as $\Anew$. Similarly we
denote $\Xiter[t]$ and $\Xiter[t+1]$ as $\Xold$ and $\Xnew$
respectively. Then the goal is to show that $\Anew$ is closer to
$\Astar$ than $\Aold$. For the purposes of our analysis, we will find
it more convenient to directly work with dot products instead of
$\ell_2$-distances (and hence avoid sign ambiguities). With this
motivation, we define the following notion of distance between two
vectors.

\begin{definition}\label{defn:dist-vector}
  For any two vectors $z,w \in \R^d$, we define the distance between
  them as follows:
\begin{align*}
  \distop{z}{w} \defas \sup_{v \perp w}
  \frac{\iprod{v}{z}}{\twonorm{v}\twonorm{z}} = \sup_{v \perp z}
  \frac{\iprod{v}{w}}{\twonorm{v}\twonorm{w}}.
\end{align*}
\end{definition}

This definition of distance suffices for our purposes due to the
following simple lemma.

\begin{lemma}
  For any two unit vectors $u,v \in \R^d$, we have
  \begin{equation*}
    \distop{u}{v} \leq \min_{z \in \{-1,1\}} \norm{zu - v}_2 \leq \sqrt{2}\distop{u}{v}.
  \end{equation*}
  \label{lemma:ell2-to-distop}
\end{lemma}
\bprf
The proof is rather straightforward. Suppose that $\iprod{u}{v}  > 0$
so that the minimum happens at $z = 1$. The other case is
identical. We can easily rewrite
\begin{align*}
  \norm{u - v}_2^2 = (2 - 2\iprod{u}{v}) \leq 2(1 - \iprod{u}{v}^2),
\end{align*}
where the final inequality follows since $0 \leq \iprod{u}{v} \leq
1$. Writing $u = \iprod{u}{v} v + v_{\perp}$, where
$\iprod{v_{\perp}}{v} = 0$, we see that:
\begin{align*}
  1 = \norm{u}_2^2 = \iprod{u}{v}^2 + \norm{v_{\perp}}^2 =
  \iprod{u}{v}^2 + \distop{u}{v}^2.
\end{align*}
Substituting this into our earlier bound, we obtain the upper
bound. For the lower bound, we note that,
\begin{align*} 
\distop{u}{v}^2 &= 1- \left(1- \frac{\|u-v\|^2}{2} \right)^2\\ &\leq
\|u-v\|^2.\end{align*} \eprf

The distance is naturally extended to matrices for our purposes by
applying it columnwise.

\begin{definition}\label{defn:dist-matrix}
  For any two $d\times r$ matrices $Z$ and $W$, we define the distance
  between them as follows:
\begin{align*}
  \distop{Z}{W} \defas \sup_{p \in [r]}
  \distop{\col{Z}{p}}{\col{W}{p}}.
\end{align*}
\end{definition}

Note that the normalization in the definition of $\distop{z}{w}$
ensures that we can apply the distance directly to the result of the
least-squares step without worrying about the effects of
normalization. This allows us to work with a closed-form expression
for $\Anew$: 
\begin{align}
  \Anew = \Ystar \pinv{\Xnew}
  &= \Astar \Xstar \pinv{\Xnew}. \label{eqn:update-A}
\end{align}

We are now in a position to prove Theorem~\ref{thm:main-altmin}.

\bprfof{Theorem~\ref{thm:main-altmin}} As an induction hypothesis, we
have $\distop{\Aold}{\Astar} < \errt$, where we recall the
definition~\eqref{eqn:errt-seq}. We will show that for every $p \in
[r]$, we will have:
\begin{align}
  \distop{\col{\Anew}{p}}{\col{\Astar}{p}} \leq\errtplus < \frac{23616 \mu_1
    s^3}{\sqrt{d}} \errt .
  \label{eqn:distop-recur}
\end{align}
This suffices to prove the theorem by appealing to
Lemma~\ref{lemma:ell2-to-distop}.

Now fix any $w \perp \col{\Astar}{p}$ such that $\twonorm{w}=1$. We
first provide a bound on $\iprod{w}{\col{\Anew}{p}}$. Now, the following holds with high probability: 
\begin{align}
  \iprod{w}{\col{\Anew}{p}}
  = \trans{w} \Astar \Xstar \col{\pinv{\Xnew}}{p}
  &\stackrel{(\zeta_1)}{\leq} \twonorm{\trans{w} \Astar}
  \twonorm{\elt{\left(\Xstar\pinv{\Xnew}\right)}{\setminus p}{p}}
  \nonumber\\
  &\stackrel{(\zeta_2)}{\leq} \mu_1 \sqrt{\frac{r}{d}} \cdot \frac{1968
    \spindex^2 \infnorm{\delX}}{\sqrt{r}} \nonumber \\
  &= \frac{17712 \mu_1 \spindex^3}{\sqrt{d}} \errt,
  \label{eqn:distop-upper}
\end{align}
where $(\zeta_1)$ follows from the fact that $\trans{w}
\col{\Astar}{p}=0$ and $(\zeta_2)$ follows from
Assumption~$(A2)$ and Lemma~\ref{lem:bound-XXstarplus}.

In order to bound $\distop{\Anew}{\Astar}$, it remains to show a lower
bound on $\norm{\Anew}_2$. This again follows using basic algebra, given our
main lemmas.
\begin{align*}
  \twonorm{\col{\Anew}{p}} = \twonorm{\Astar \Xstar \col{\pinv{\Xnew}}{p}}
  &= \twonorm{\Astar \left(\Xnew - \delX\right) \col{\pinv{\Xnew}}{p}} \\
  &\stackrel{(\zeta_1)}{=} \twonorm{\col{\Astar}{p} - \Astar \delX
    \col{\pinv{\Xnew}}{p}} \\
  &\geq \twonorm{\col{\Astar}{p}} - \twonorm{\Astar \col{\left(\delX
      \pinv{\Xnew}\right)}{p}},
\end{align*}
where $(\zeta_1)$ follows from the fact that $\Xnew \pinv{\Xnew} =
\eye$. We decompose the second term into diagonal and off-diagonal
terms of $\delX\pinv{\Xnew}$, followed by triangle inequality and
obtain:
\begin{align*}
  \twonorm{\col{\Anew}{p}} &\geq 1 - \twonorm{\col{\Astar}{p}
    \elt{\left(\delX \pinv{\Xnew}\right)}{p}{p}
    + \col{\Astar}{\setminus p} \elt{\left(\delX
      \pinv{\Xnew}\right)}{\setminus p}{p}} \\
  &\geq 1 - \twonorm{\col{\Astar}{p}} \abs{\elt{\left(\delX
      \pinv{\Xnew}\right)}{p}{p}}
  - \twonorm{\col{\Astar}{\setminus p}} \twonorm{\elt{\left(\delX
      \pinv{\Xnew}\right)}{\setminus p}{p}} \\
  &\geq 1 - 1 \cdot \twonorm{\delX
    \trans{\Xnew}\inv{\left(\Xnew\trans{\Xnew}\right)}}
  - \twonorm{\col{\Astar}{\setminus p}} \twonorm{\elt{\left(\delX
      \pinv{\Xnew}\right)}{\setminus p}{p}} \\
  &\geq 1 - \underbrace{\twonorm{\delX} \twonorm{\trans{\Xnew}}
  \twonorm{\inv{\left(\Xnew\trans{\Xnew}\right)}} }_{\term_1}
  - \underbrace{\twonorm{\col{\Astar}{\setminus p}}
    \twonorm{\elt{\left(\delX \pinv{\Xnew}\right)}{\setminus
        p}{p}}}_{\term_2}
\end{align*}
It remains to control $\term_1$ and $\term_2$ at an appropriate
level. We start from $\term_1$. Note that $\twonorm{\delX}$ is bounded
by Lemmas~\ref{lem:update-X} and \ref{lem:W-spectralnorm}, while
$\twonorm{\trans{\Xnew}}$ is controlled by
Lemma~\ref{lem:Xnew-spectralnorm} in the appendix (recall
$\norm{\delX}_\infty \leq 1/(64s)$). Invoking
Lemma~\ref{lem:Xnew-covariance-spectralnorm} to control
$\twonorm{\inv{(X\trans{X})}}$, we obtain the following bound on
$\term_1$ with probability at least $1- r\exp\left(-\frac{C
  n}{rM^2s}\right)$:
\begin{equation*}
  \term_1 \leq 18 \errt \spindex^2 \sqrt{\frac{n}{r}} \cdot 3
  \spindex\sqrt{\frac{n}{r}} \cdot \frac{8r}{sn} = 432 \spindex^2
  \errt.
\end{equation*}
The second term $\term_2$ is directly controlled by
Lemma~\ref{lem:bound-XXstarplus}, yielding the following (with probability at least 
\mbox{$1 - r\exp\left(-\frac{C
    n}{rM^2s}\right) -\exp\left(-Cn/r^2\right)$}): 
\begin{equation*}
  \term_2 \leq \mu_1 \sqrt{\frac{r}{d}} \frac{1968 \spindex^3
    \errt}{\sqrt{r}}.
\end{equation*}
Putting all the terms together, we get:
\begin{align}
  \twonorm{\col{\Anew}{p}} \geq 1 - 9\spindex^2\left(48+\frac{1968
    \spindex \mu_1}{\sqrt{d}}\right)\errt \geq \frac{3}{4},
  \label{eqn:distop-lower}
\end{align}
where the inequality follows since $9\spindex^2\left(48+\frac{1968
  \spindex \mu_1}{\sqrt{d}}\right)\errt \leq
9\spindex^2\left(48+\frac{1968 \spindex
  \mu_1}{\sqrt{d}}\right)\errt[0] \leq 1/4$ by our assumption $(A6)$
on $\errt[0]$. Combining the bounds~\eqref{eqn:distop-upper}
and~\eqref{eqn:distop-lower} yields the desired
recursion~\eqref{eqn:distop-recur}. Appealing to
Lemma~\ref{lemma:ell2-to-distop} along with our setting of
$\errt$~\eqref{eqn:errt-seq} completes the proof of the
claim~\eqref{eqn:distop-recur}. Finally note that the error
probability in the theorem is obtained by using the fact that $M \geq
1$, and that the failure probability is purely incurred from the
structure of the non-zero entries of $\Xstar$, so that it is incurred
only once and not at each round. This avoids the need of a union bound
over all the rounds, yielding the result.  \eprfof



\section{Experiments}
\begin{figure*}[t]
  \centering
  \begin{tabular}{ccc}
\hspace*{-10pt}\includegraphics[width=.35\textwidth]{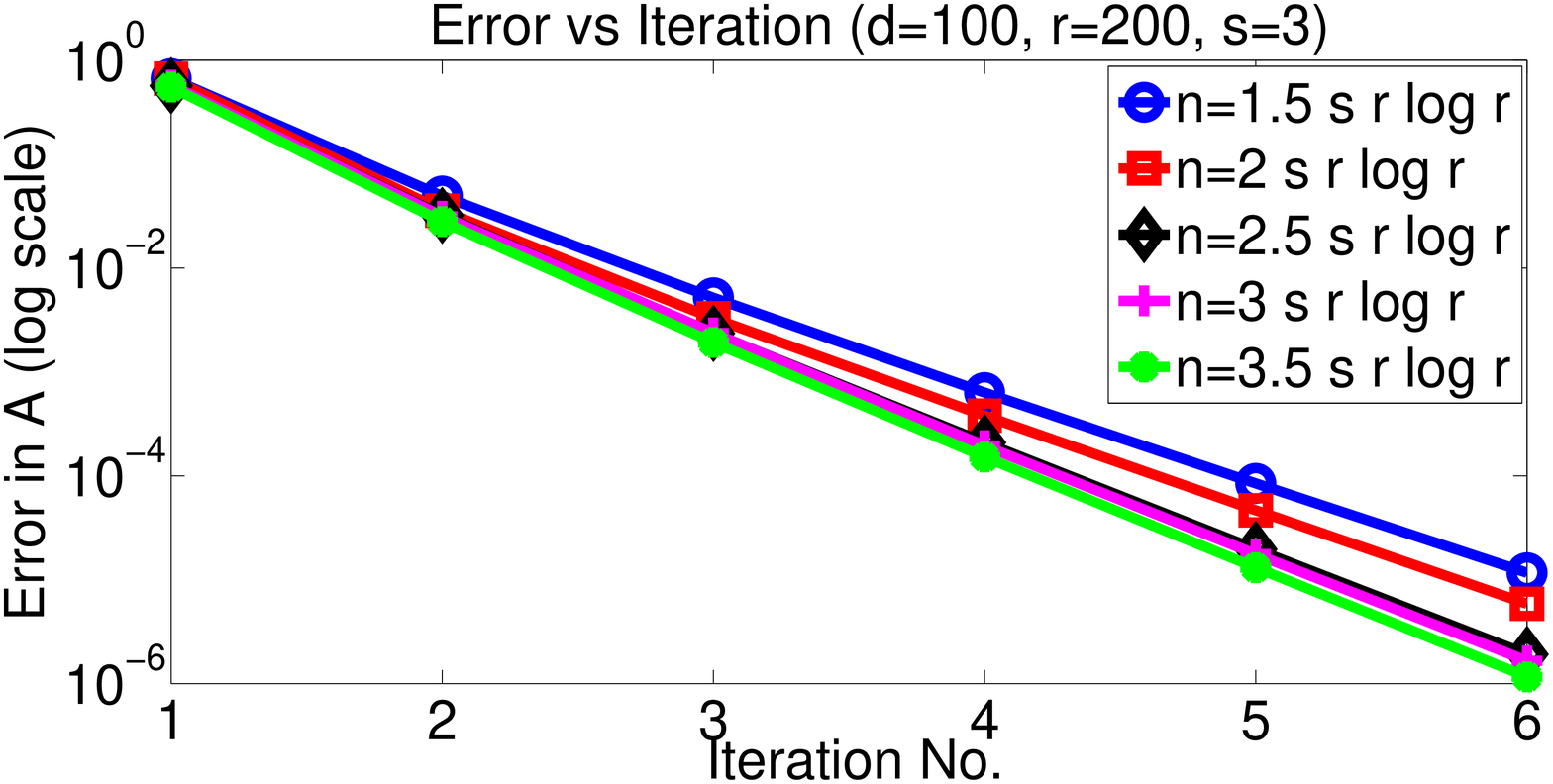}&
\hspace*{-20pt}\includegraphics[width=.35\textwidth]{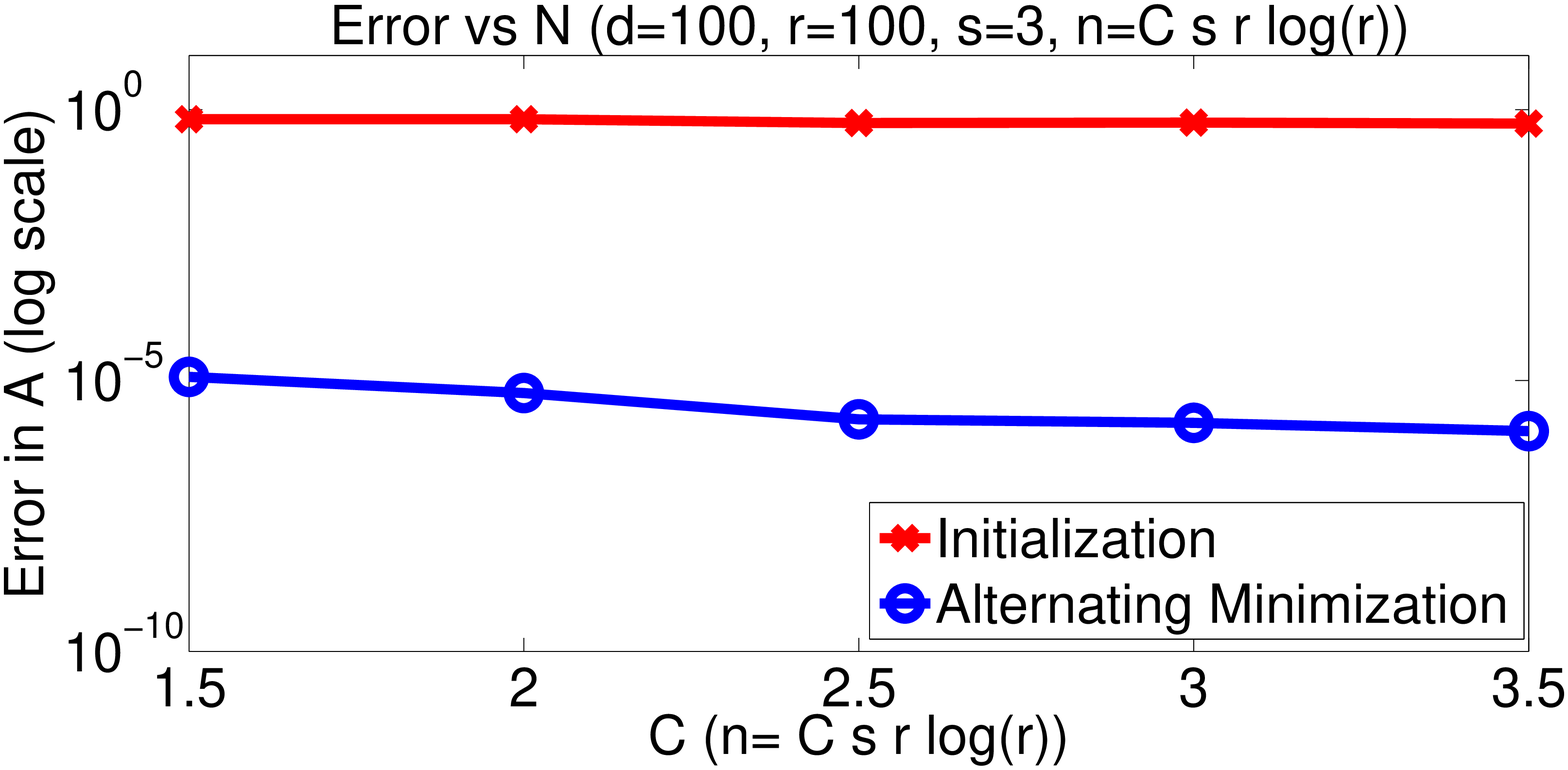}&
\hspace*{-25pt}\includegraphics[width=.35\textwidth]{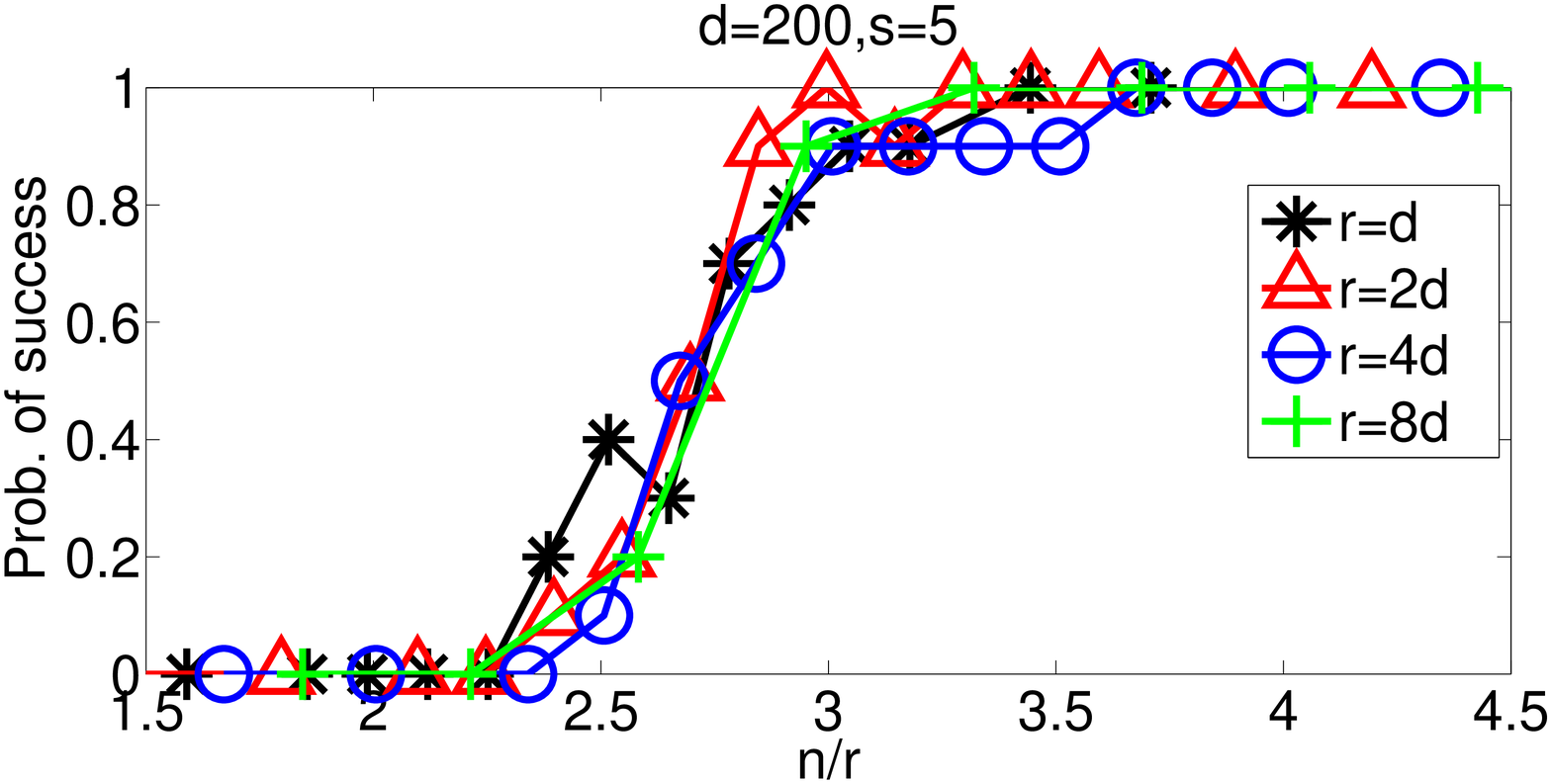}\\
(a)&(b)&(c)
  \end{tabular}\vspace*{-10pt}
\caption{(a): Average error after each alternating minimization
  step of Algorithm~\ref{algo:altmin} on log-scale. (b): Average error
  after the initialization procedure (Algorithm~1 of \cite{AgarwalAN13})
  and after $5$ alternating minimization steps of
  Algorithm~\ref{algo:altmin}. (c): Sample complexity requirement of
  the alternating minimization algorithm. For ease of experiments, we
  initialize the dictionary using a random perturbation of the true
  dictionary rather than using Algorithm~1 of \cite{AgarwalAN13} which
  should in fact give better initial point with smaller error.}
\label{fig:main}
\end{figure*}
Alternating minimization/descent approaches have been widely used for
dictionary learning and several existing works show effectiveness of
these methods on real-world/synthetic datasets \cite{KKG2013,
  TRS:dict2013}. Hence, instead of replicating those results, in this
section we focus on illustrating the following three key properties of
our algorithms via experiments in a controlled setting: a) advantage
of alternating minimization over one-shot initialization, b) linear
convergence of alternating minimization, c) sample complexity of
alternating minimization.

\textbf{Data generation model}: Each entry of the dictionary matrix
$A$ is chosen i.i.d. from $\N(0,1/\sqrt{d})$. Note that, random
Gaussian matrices are known to satisfy incoherence and the spectral
norm bound \cite{Vershynin2010}. The support of each column of $X$ was
chosen independently and uniformly from the set of all
$\spindex$-subsets of $[r]$. Similarly, each non-zero element of $X$
was chosen independently from the uniform distribution on $[-2,-1]\cup
[1,2]$. We use the GraDeS algorithm of \cite{GargK09} to solve the
sparse recovery step, as it is faster than lasso. We measure error in
the recovery of dictionary by $error(A)=\max_i \sqrt{1-\frac{\langle
    A_i, A_i^*\rangle^2}{\|A_i\|_2^2\|A^*_i\|_2^2}}$.  The first two
plots are for a typical run and the third plot averages over $10$
runs.  The implementation is in Matlab.

{\bf Linear convergence}: In the first set of experiments, we fixed
$d=100$, $r=200$ and measured error after each step of our algorithm
for increasing values of $n$. Figure~\ref{fig:main} (a) plots error
observed after each iteration of alternating minimization; the first
data point refers to the error incurred by the initialization method
(Algorithm~1 of \cite{AgarwalAN13}).
As expected due to Theorem~\ref{thm:main-altmin}, we observe a
geometric decay in the error.

{\bf One-shot vs iterative algorithm}: It is conceivable that a good
initialization procedure itself is sufficient to obtain an estimate
of the dictionary upto reasonable accuracy, without recourse to the
alternating minimization procedure of Algorithm~\ref{algo:altmin}.
Figure~\ref{fig:main}(b) shows that this is not the case.  The figure
plots the error in recovery vs the number of samples used for both
Algorithm 1 of Agarwal et al.~\cite{AgarwalAN13} and
Algorithm~\ref{algo:altmin}. It is clear that the recovery error of
the alternating minimization procedure is significantly smaller than
that of the initialization procedure. For example, for $n=2.5 s r \log
r$ with $s=3, r=200, d=100$, initialization incurs error of $.56$
while alternating minimization incurs error of $10^{-6}$.  Note
however that the recovery accuracy of the initialization procedure is
non-trivial and also crucial to the success of alternating
minimization- a random vector in $\R^d$ would give an error of
$1-\frac{1}{d}=0.99$ (since the inner product is concentrated around
$1/\sqrt{d}$), where as the error after initialization procedure is
$\approx 0.55$.

\textbf{Sample complexity}: Finally, we study sample complexity
requirement of the alternating minimization algorithm which is $n =
\order{r^2 \log r}$ according to Theorem~\ref{thm:main-altmin},
assuming good enough initialization.  Figure~\ref{fig:main}(c)
suggests that in fact only $\order{r}$ samples are sufficient for
success of alternating minimization. The figure plots the probability
of success with respect to $\frac{n}{r}$ for various values of $r$.
A trial is said to succeed if at the end of $25$ iterations, the error
is smaller than $10^{-6}$. Since we focus only on the sample
complexity of alternating minimization, we use a faster initialization
procedure: we initialize the dictionary by randomly perturbing the
true dictionary as $ A(0) = \Astar + Z,$ where each element of $Z$ is
an $\N(0,0.5/\sqrt{d})$ random variable.
Figure~\ref{fig:main} (c) shows that the success probability
transitions at nearly the same value for various values of $r$,
suggesting that the sample complexity of the alternating minimization
procedure in this regime of $r = \order{d}$ is just $O(r)$.
\section{Conclusions}

In this paper we provide the first analysis for the local linear
convergence of the popular alternating minimization heuristic commonly
used for solving dictionary learning problems in practice. Combined
with some recent results, this also provides an efficient method for
global and exact recovery of the unknown overcomplete dictionary under
favorable assumptions. The results are of interest from both
theoretical and practical standpoints. From a theoretical standpoint,
this is one of the very few results that provides guarantees on a
dictionary learned using an efficient algorithm, and one of the first
for the overcomplete setting. From a practical standpoint, there is a
tremendous interest in the problem, and we believe that an
understanding of the theoretical properties of existing methods is
critical in designing better methods. Indeed, our work provides some
such hints towards designing a better algorithm. For instance, the
sparse recovery step in our method decodes the coefficients
individually for each sample. We believe that a better method can be
designed by jointly decoding all the samples, which allows one to
force consistency across samples (for instance, in our random
coefficient model, the number of samples per dictionary element is
also controlled in addition to the number of dictionary elements per
sample). 

More interestingly, our work extends a growing body of recent
literature on analysis of alternating minimization methods for a
variety of non-convex factorization problems~\cite{JainNeSa2013,
  NetrapalliJaSa2013}, where global in addition to local results are
being established with appropriate initialization strategies. Of
course, results on alternating minimization go much further back, even
in non-convex optimization to Csiszar's seminal works (see, e.g. the
recent tutorial~\cite{CsiszarSh2004} for an overview), as well as in
convex minimization and projection problems. However, the recent work
has been largely motivated by applications of non-convex optimization
arising in machine learning. We believe that the emergence of these
newer results indicates the possibility of a more general theory of
alternating optimization procedures for a broad class of
factorization-style non-convex problems, and should be an exciting
question for future research

\bibliography{../bib}
\bibliographystyle{abbrv}

\newpage

\appendix
\section{Proofs for alternating minimization}\label{app:altmin}
In this section, we will present our proof for the results on
alternating minimization. We present the proofs for
Theorem~\ref{thm:main-altmin} and the other main lemmas in
Section~\ref{sec:altmin-mainproofs}. In Section \ref{sec:technical},
we present the auxiliary lemmas and their proofs.

\subsection{Proofs of main lemmas}\label{sec:altmin-mainproofs}

For reader's convenience, we recall Lemmas~\ref{lem:update-X},
~\ref{lem:W-spectralnorm} and \ref{lem:bound-XXstarplus} from
Section~\ref{sec:overview} along with their proofs. The more technical
lemmas are deferred to the next section.

We first recall some notation and define additional abbreviations before
proving the lemmas.  Denote $\Xoip=\dip\Mip, \ \forall 1\leq p\leq
r,\ \forall 1\leq i\leq n$ where $\dip=1$ if $p \in \supp(\Xstar_i)$
and $0$ otherwise and $\Mip$ are i.i.d. random variables with
$\expec{\Mip}=\mu$ and
$\expec{(\Mip)^2}=\sigma^2+\mu^2$. Assumption~$(A3)$ gives us:
\begin{enumerate}
  \item	$\mu^2+\sigma^2 = 1$, and
  \item	$|\Mip|\leq M$ a.s.
\end{enumerate}

\begin{replemma}{lem:update-X}[Error in sparse recovery]
  Let $\delX \defas \Xold - \Xstar$. Assume that $2\mu_0s/\sqrt{d}
  \leq 0.1$ and $\sqrt{s\errt} \leq 0.1$ Then, we have:
\begin{enumerate}
  \item	$\nzset(\delX) \subseteq \nzset(\Xstar)$.
  \item $\infnorm{\delX} \leq 9s \cdot \distop{\Aold}{\Astar} \leq
    9s\errt$.
\end{enumerate}
\end{replemma}

\bprf In order to establish the lemma, we
use a result of Candes regarding the lasso estimator with
deterministic noise for the recovery procedure:

\begin{equation}
  \widehat{x}_i = \arg\min_{x \in \R^r} \norm{x}_1 \quad \mbox {such
    that}, \quad \norm{Y_i - A x}_2 \leq \epsilon.
  \label{eqn:lasso}
\end{equation}

\begin{theorem}[Theorem 1.2 from~\citet{Candes2008}]
  Suppose $Y_i = A x_i + z_i$, where $x_i$ is
  $s$-sparse and $\norm{z_i}_2 \leq \epsilon$. Assume further that
  $\delta_{2s} \leq \sqrt{2}-1$. Then the solution to
  Equation~\eqref{eqn:lasso} obeys the following, for a universal
  constant $C_1$,
  \begin{equation*}
    \norm{\widehat{x}_i - x_i}_2 \leq C_1\epsilon
  \end{equation*}
  In particular, $C_1 = 8.5$ suffices for $\delta_{2s} \leq 0.2$.
  \label{thm:candes}
\end{theorem}

In order to apply the theorem, we need to demonstrate that the RIP
condition holds on $\Aold$. Consider any $2\spindex$-sparse subset $S$
of $[r]$. We have:

\begin{align*}
  \singmin(\Aold_S) &\geq \singmin(\Astar_S) - \norm{\Astar_S -
    \Aold_S}_2 \stackrel{(\zeta_1)}{\geq} 1 - \delta_{2s}
    - \frob{\Astar_S - \Aold_S} \quad \mbox{ and,}
  \\ \singmax(\Aold_S) &\leq \singmax(\Astar_S) + \norm{\Astar_S - \Aold_S}_2
  \stackrel{(\zeta_2)}{\leq} 1 + \delta_{2s} +
  \frob{\Astar_S - \Aold_S},
\end{align*}
where $\zeta_1$ and $\zeta_2$ follow from
Assumption~$(A1)$. Recalling the assumption $\sqrt{s\errt} < 0.1$,
and that $\delta_{2s} < 0.1$,
we see that the maximum and minimum singular values of $\Aold_S$ are
at least $4/5$ and at most $6/5$ respectively. Appealing to
Theorem~\ref{thm:candes}, we see that this guarantees
$\norm{\delX_i}_2 \leq 9s\errt$. Since this is also an infinity norm
error bound, we obtain the second part of the lemma. The proof of the
first part is further implied by the choice of our threshold at a
level of $9s\errt$, which ensures that any non-zero element in $\Xnew$
has $\abs{\elt{\Xstar}{i}{p}} \geq 0$ (since we would have
$\abs{\elt{\Xnew}{i}{p}} \leq 9s\errt$ by our infinity norm bound
otherwise).  
\eprf

We now move on to the proof of Lemma~\ref{lem:W-spectralnorm}. We
point out that the lemma applies uniformly to all matrices $W$
satisfying $\supp(W) \subseteq \supp(\Xstar)$, irrespective of the
values of these entries. This might be surprising at first, but is a
rather straight forward consequence of random matrix concentration
theory.

\begin{replemma}{lem:W-spectralnorm}
For every $r \times n$ matrix $W$ s.t. $\supp(W)\subseteq
\supp(\Xstar)$, we have (w.p. $\geq 1-r\exp\left(-\frac{C
  n}{rs}\right)$):
\begin{align*}
  \twonorm{W} \leq 2\|W\|_\infty\sqrt{\frac{s^2n}{r}}.
\end{align*}
\end{replemma}

\bprf
Since the support of $W$ is a subset of the support of $\Xstar$,
$W^p_i=\dip W^p_i$. Now,

\begin{align*}
\|W\|_2 &= \max_{u,v \|u\|_2=1, \norm{v}_2=1}\sum_{ip}W^p_iu^iv^p =
\max_{u,v \|u\|_2=1, \norm{v}_2=1}\sum_{ip}\dip W^p_iu^iv^p\\
&\leq \|W\|_\infty\cdot \max_{u,v \|u\|_2=1, \norm{v}_2=1}\cdot
\sum_{ip} \dip u^iv^p,
\end{align*}
where the inequality holds since the maximum inner product over the
all pairs $(u,v)$ from the unit sphere is larger than that over pairs
with $\row{u}{i}\row{v}{p} \geq 0$ for all $i,p$. Note that the last
expression is equal to $\norm{W}_\infty \trans{u}\spmat v$, where we
use $\spmat$ to denote the matrix with the non-zero pattern of the
matrix $\Xstar$. It suffices to control the operator norm of this
matrix for proving the lemma. This can indeed be done by applying
Lemmas~\ref{lem:expectedcovariance}
and~\ref{lem:empiricalcovariance-concentration} with $\mu = M = 1$ and
$\sigma = 0$. Doing so, yields with probability at least $
1-r\exp\left(-\frac{C n}{rs}\right)$

\begin{align*}
  \twonorm{W} \leq 2\|W\|_\infty\sqrt{\frac{s^2n}{r}},
\end{align*}
which completes the proof.
\eprf

We now finally prove Lemma~\ref{lem:bound-XXstarplus}, which is our
main lemma on the structure of $\Xstar\pinv{X}$. Specifically, the
lemma will show how to control the off-diagonal elements of this
matrix carefully.

\begin{replemma}{lem:bound-XXstarplus}[Off-diagonal error bound]
Suppose $\infnorm{\delX} < \frac{1}{288 \spindex}$. Then with
probability at least \mbox{$1 -
  r\exp\left(-\frac{C n}{rM^2s}\right)
  -r \exp\left(-C n/r^2\right)$}, we have uniformly for every $p \in
[r]$,
\begin{align*}
  \twonorm{\elt{\left(\delX \pinv{\Xnew}\right)}{\setminus p}{p}} =
  \twonorm{\elt{\left(\Xstar \pinv{\Xnew}\right)}{\setminus p}{p}}
  \leq \frac{1968 \spindex^2 \infnorm{\delX} }{\sqrt{r}}.
\end{align*}
\end{replemma}

\bprf
For simplicity, we will prove the statement for $p=1$.
We first relate $\Xstar \pinv{\Xnew}$ to $\delX \pinv{\Xnew}$.

\begin{align*}
  \elt{\left(\Xstar \pinv{\Xnew}\right)}{\setminus 1}{1}
  &=   {\elt{\left(\left(\Xstar-\Xnew\right)
      \pinv{\Xnew}\right)}{\setminus 1}{1}} \\
  &=   -{\elt{\left(\delX \pinv{\Xnew}\right)}{\setminus 1}{1}} \\
  &=   -{\elt{\left(\delX \trans{\Xnew} \inv{\left(\Xnew
        \trans{\Xnew}\right)}\right)}{\setminus 1}{1}},
\end{align*}

where the first step follows from the fact that $\Xnew \pinv{\Xnew} =
\eye$.  This proves the first part of the lemma.  We now expand the
above as follows:

\begin{align*}
  \elt{\left(\delX \trans{\Xnew} \inv{\left(\Xnew
      \trans{\Xnew}\right)}\right)}{\setminus 1}{1}
  &= \elt{\left(\delX \trans{\Xnew}\right)}{\setminus 1}{1}
  \elt{\left(\inv{\left(\Xnew \trans{\Xnew}\right)}\right)}{1}{1}
  + \elt{\left(\delX \trans{\Xnew}\right)}{\setminus 1}{\setminus 1}
  \elt{\left(\inv{\left(\Xnew \trans{\Xnew}\right)}\right)}{\setminus
    1}{1}.
\end{align*}
Using triangle inequality, we have:
\begin{align}
  \twonorm{\elt{\left(\delX \trans{\Xnew} \inv{\left(\Xnew
        \trans{\Xnew}\right)}\right)}{\setminus 1}{1}}
  &\leq \underbrace{\abs{\elt{\left(\inv{\left(\Xnew
          \trans{\Xnew}\right)}\right)}{1}{1}}}_{\term_1}
  \underbrace{\twonorm{\elt{\left(\delX
        \trans{\Xnew}\right)}{\setminus 1}{1}}}_{\term_2} \nonumber \\
  &\qquad \qquad + \underbrace{\twonorm{\elt{\left(\delX
        \trans{\Xnew}\right)}{\setminus 1}{\setminus 1}}}_{\term_3}
  \underbrace{\twonorm{\elt{\left(\inv{\left(\Xnew
          \trans{\Xnew}\right)}\right)}{\setminus 1}{1}}}_{\term_4}.
  \label{eqn:bound-XXstarplus-triangle}
\end{align}

We now bound each of the above four quantities. We can easily bound
$\term_1$ via a spectral norm bound on
$\inv{(\Xnew\trans{\Xnew})}$. Doing so, we obtain with probability at
least $1-r\exp(-\frac{C n }{r M^2s})$

\begin{align}
  \term_1 = \abs{\elt{\left(\inv{\left(\Xnew
        \trans{\Xnew}\right)}\right)}{1}{1}} \leq
  \twonorm{\inv{\left(\Xnew \trans{\Xnew}\right)}}
  &\stackrel{(\zeta_1)}{\leq} \frac{8r}{n \spindex},
	\label{eqn:bound-XXstarplus-I}
\end{align}

where $(\zeta_1)$ follows from
Lemma~\ref{lem:empiricalcovariance-concentration}. To bound $\term_2$,
we use Lemma~\ref{lem:2-intersection} and obtain with probability at
least \mbox{$1 - r\exp\left(-\frac{C n }{r^2}\right) -
  r\exp\left(-\frac{C n}{rM^2s}\right)$}

\begin{align}
  \term_2 = \twonorm{\elt{\left(\delX \trans{\Xnew}\right)}{\setminus
      1}{1}} &\leq \frac{6 \infnorm{\delX} s^2 n
  }{r^{\frac{3}{2}}}, \label{eqn:bound-XXstarplus-II}
\end{align}
where we recall the assumption $\norm{\delX}_\infty \leq 1/(64s)$.
We now bound $\term_3$ as follows

\begin{align}
  \term_3 = \twonorm{\elt{\left(\delX \trans{\Xnew}\right)}{\setminus
      1}{\setminus 1}} \leq \twonorm{\elt{\left(\delX
      \right)}{\setminus 1}{\setminus 1}} \twonorm{\elt{\left(\Xnew
      \right)}{\setminus 1}{\setminus 1}} &\stackrel{(\zeta_1)}{\leq}
  2 \infnorm{\delX} \spindex \sqrt{\frac{n}{r}} \cdot 2 (1 +
  \infnorm{\delX} ) \spindex \sqrt{\frac{n}{r}}\nonumber \\
  &< \frac{ 6 \infnorm{\delX} \spindex^2 n
  }{r},\label{eqn:bound-XXstarplus-III}
\end{align}

where $(\zeta_1)$ follows from Lemmas~\ref{lem:W-spectralnorm} and
\ref{lem:Xnew-spectralnorm} (since $\nzset(\delX) \subseteq
\nzset(\Xnew) \cup \nzset(\Xstar) = \nzset(\Xstar)$). Finally, to
bound $\term_4$, we start by noting the following block decomposition
of the matrix $\Xnew\trans{\Xnew}$

\begin{equation*}
  X\trans{X} = \left[\begin{array}{cc} \row{X}{1}\trans{(\row{X}{1})}
      & \row{X}{1}\trans{(\row{X}{\setminus
          1})}\\ \row{X}{\setminus 1} \trans{\row{X}{1}} &
      \row{X}{\setminus 1}\trans{(\row{X}{\setminus 1})} \end{array}
    \right].
\end{equation*}
Given this block-structure, we can now invoke
Lemma~\ref{lem:Schurcomplement} (Schur complement lemma) to obtain

\begin{align*}
\elt{\left(\inv{\left(\Xnew \trans{\Xnew}\right)}\right)}{\setminus
  1}{1} = -\frac{1}{\row{\Xnew}{1}\trans{\left(\row{\Xnew}{1}\right)}}
B \row{\Xnew}{\setminus 1}\trans{\left(\row{\Xnew}{1}\right)},
\end{align*}
where,

\begin{align}\label{eqn:Xinv-schur}
B \defas \elt{\left(\inv{\left(\Xnew
    \trans{\Xnew}\right)}\right)}{\setminus 1}{\setminus 1}
.
\end{align}
Here we recall that $\inv{B}$ is the Schur complement of
$\col{X}{1}\trans{\col{X}{1}}$. Using Lemma~\ref{lem:2-intersection}
and Equation~\ref{eqn:element-lb-xxtrans} we have with probability at
least \mbox{$1 - r\exp\left(-\frac{C n}{rM^2s}\right) -\exp\left(-C
  n/r^2\right)$}

\begin{align}
  \twonorm{\elt{\left(\inv{\left(\Xnew
        \trans{\Xnew}\right)}\right)}{\setminus 1}{1}} \leq
  \frac{1}{\abs{\row{\Xnew}{1}\trans{\left(\row{\Xnew}{1}\right)}}}
  \twonorm{B} \twonorm{\row{\Xnew}{\setminus
      1}\trans{\left(\row{\Xnew}{1}\right)}} \leq \frac{8r}{sn} \cdot
  \twonorm{B} \cdot \frac{5s^2 n}{r^{\frac{3}{2}}} =
  \frac{40s}{\sqrt{r}} \twonorm{B}.
  \label{eqn:bound-XXstarplus-IV-I}
\end{align}

Using the expression~\eqref{eqn:Xinv-schur} and the lower bound on
$\singmin(X)$ from Lemma~\ref{lem:Xnew-spectralnorm}, we also have the
following bound for $\twonorm{B}$ with probability at least \mbox{$1 -
  r\exp\left(-\frac{Cn}{rM^2s}\right)$}:

\begin{align*}
  \twonorm{B} = \twonorm{\elt{\left(\inv{\left(\Xnew
        \trans{\Xnew}\right)}\right)}{\setminus 1}{\setminus 1}} &\leq
  \twonorm{\inv{\left(\Xnew \trans{\Xnew}\right)}} \leq \frac{8r}{n
    \spindex}.
\end{align*}
Plugging the above into \eqref{eqn:bound-XXstarplus-IV-I}, gives us:

\begin{align}
  \twonorm{\elt{\left(\inv{\left(\Xnew
        \trans{\Xnew}\right)}\right)}{\setminus 1}{1}} &\leq
  \frac{40 \spindex}{\sqrt{r}} \cdot \frac{8r}{n \spindex} \leq
  \frac{320\sqrt{r}}{n}. \label{eqn:bound-XXstarplus-IV}
\end{align}

Combining \eqref{eqn:bound-XXstarplus-I},
\eqref{eqn:bound-XXstarplus-II}, \eqref{eqn:bound-XXstarplus-III} and
\eqref{eqn:bound-XXstarplus-IV}, we obtain with probability at least\\
\mbox{$1 - r\exp\left(-\frac{C
    n}{rM^2s}\right) -\exp\left(-Cn/r^2\right)$}

\begin{align*}
  \twonorm{\elt{\left(\Xnew \pinv{\Xstar}\right)}{\setminus p}{p}} &\leq
  \frac{48 \infnorm{\delX}\spindex}{\sqrt{r}} + \frac{1920
    \infnorm{\delX}\spindex^2}{\sqrt{r}}\\
  &\leq \frac{1968 \spindex^2 \infnorm{\delX} }{\sqrt{r}}.
\end{align*}
\eprf

\subsection{Main Technical Lemmas}
\label{sec:technical}

In this section, we state and prove the main technical lemmas used in
our results.

\begin{lemma}\label{lem:expectedcovariance}
We have:
\begin{align*}
  \Sigma \defas \expec{\col{\Xstar}{i}
    \trans{\col{\Xstar}{i}}}=\left(\frac{s}{r} -
  \frac{s(s-1)\mu^2}{r(r-1)} \right) \eye +
  \frac{s(s-1)\mu^2}{r(r-1)}\ones \trans{\ones}.
\end{align*}
\end{lemma}
\bprf

Note that, $\dip, 1\leq p\leq r$ all have same distribution. Hence, by
symmetry and linearity of expectation,
$\expec{\dip}=\frac{1}{r}\expec{\sum_{q=1}^r
  \diq}=\frac{s}{r}$. Similarly,
$\expec{(\dip)^2}=\frac{1}{r}\expec{\sum_{q=1}^r
  (\diq)^2}=\frac{s}{r}$. Also, $\expec{(\sum_{q=1}^r
  (\diq))^2}=\expec{\sum_{p,q}\dip\diq}=r\expec{(\dip)^2}+(r^2-r)\expec{\dip\diq}$. Hence,
$\expec{\dip\diq}=\frac{s(s-1)}{r(r-1)}$.

Now, recall that $\Xoip=\dip\Mip$. Now,  we first consider diagonal
terms of $\Sigma$:

\begin{equation}
  \Sigma_p^p = \expec{(\Xoip)^2} = \expec{(\dip)^2}\expec{(\Mip)^2} =
  \frac{s}{r}(\mu^2+\sigma^2) = \frac{s}{r}.
  \label{eq:sigmaii}
\end{equation}
Similarly, using independence of $\Xoip$ and $\elt{\Xstar}{j}{p}$,
off-diagonal terms of $\Sigma$ are given by:
\begin{equation}
  \label{eq:sigmaij}
  \Sigma_p^q = \expec{\dip\diq}\expec{\Mip}\expec{\Miq} =
  \frac{s(s-1)}{r(r-1)}\mu^2.
\end{equation}
Lemma now follows by using \eqref{eq:sigmaii} and \eqref{eq:sigmaij}.
\eprf

In particular, two consequences of the lemma which will be
particularly useful are about the extreme singular values of
$\Sigma$. Recalling that $2s \leq r$ and $\mu^2 \leq 1$ by assumption,
we obtain

\begin{equation}
  \singmin(\Sigma) \geq \frac{s}{2r}, \quad \mbox{and} \quad
  \singmax(\Sigma) \leq \frac{2s^2}{r}.
  \label{eqn:extreme-cov-pop}
\end{equation}

We next establish some results on the spectrum of the empirical
covariance matrix, using a standard result from random matrix
theory. For convenience of the reader, we recall the following theorem
from \cite{Vershynin2010}.

\begin{theorem}[Restatement of Theorem~5.44 from \cite{Vershynin2010}]\label{thm:rand-matrix-spectralnorm-vershynin}
Consider a $r \times n$ matrix ${W}$ where each column ${w_i}$ of
${W}$ is an independent random vector with covariance matrix
${\Sigma}$. Suppose further that $\twonorm{w_i} \leq
\sqrt{u}$ a.s. for all $i$. Then for any $t \geq 0$, the following inequality
holds with probability at least $1 - r \exp\left(-ct^2\right)$:
\begin{align*}
  \twonorm{\frac{1}{n} W W^T - \Sigma} \leq
  \max\left(\twonorm{\Sigma}^{1/2}\gamma,\gamma^2\right) \mbox{ where
  } \gamma = t \sqrt{\frac{u}{n}}.
\end{align*}
Here $c>0$ is an absolute numerical constant. In particular, this inequality yields:
\begin{align*}
  \twonorm{W} \leq \twonorm{\Sigma}^\frac{1}{2}
  \sqrt{n} + t \sqrt{u}.
\end{align*}
\end{theorem}

Using the theorem, we can establish the following results on
concentration of empirical covariance matrices. Hereafter, $C$ will be
a universal constant that can change from line to line.

\begin{lemma}\label{lem:empiricalcovariance-concentration}
There exists a universal constant $C$ such that with probability at
least $ 1-r\exp(-\frac{C \delta^2 n \spindex}{r M^2})$, we have:
\begin{align*}
  \twonorm{\frac{1}{n}\Xstar \trans{\Xstar} - \Sigma} \leq
  \max\left(\sqrt{2}\delta, \delta^2\right) \frac{s^2}{r}.
\end{align*}
In particular, with probability at least $1-r\exp(-\frac{C n}{r
  M^2s})$, we have the bounds
\begin{align*}
   \twonorm{\Xstar}\leq 2\sqrt{\frac{ns^2}{r}} \quad \mbox{and} \quad
   \singmin(\Xstar) \geq \sqrt{\frac{n\spindex}{4r}}.
\end{align*}
\end{lemma}

\bprf

Note that, $\|\Xstar_i\|_2\leq \sqrt{s}M$. Also, $\|\Sigma\|_2\leq
\frac{s}{r}+\frac{s(s-1)\mu^2}{r-1}\leq \frac{2s^2}{r}$.  Using
Theorem~\ref{thm:rand-matrix-spectralnorm-vershynin} with $t =
\delta\sqrt{\frac{n\spindex}{rM^2}}$, we obtain:

$$\twonorm{\frac{1}{n}\Xstar \trans{\Xstar} - \Sigma} \leq
\max\left(\sqrt{2}\delta, \delta^2\right) \frac{s^2}{r},$$
w.p. greater than $1 - r \exp\left(-\frac{C \delta^2 n \spindex}{r
  M^2}\right)$. In order to obtain the second part, we apply the first
part of the lemma with $\delta = 1/4\sqrt{2}s$ as well as
Lemma~\ref{lem:expectedcovariance} to bound the largest and smallest
singular values of $X\trans{X}/n$. Taking square roots completes the
proof.

\eprf

The next lemma we state is a specialization of
Lemma~\ref{lem:W-spectralnorm} to obtain bounds on the spectral norm
of our iterates $\Xnew$.

\begin{lemma}\label{lem:Xnew-spectralnorm}
With probability at least $1-r\exp\left(-\frac{C n}{rs}\right)-
r\exp\left(-\frac{C n}{rM^2s}\right)$, for every $r \times n$ matrix
$\Xnew$ s.t. $\supp(\Xnew)\subseteq \supp(\Xstar)$, we have:
\begin{align*}
  \twonorm{\Xnew} \leq 2\cdot \left(1 + \infnorm{\Xnew -
    \Xstar}\right) \cdot \spindex \sqrt{\frac{n}{r}}.
\end{align*}
\end{lemma}

\bprf

Let $X=\Xstar+E_{\Xstar}$ where $\supp(E_{\Xstar})\subseteq
\supp(\Xstar)$. Hence, $\twonorm{\Xnew} \leq
\twonorm{\Xstar}+\twonorm{\Xnew-\Xstar}$.  Lemma follows directly
using Lemma~\ref{lem:empiricalcovariance-concentration} and
Lemma~\ref{lem:W-spectralnorm}.
\eprf

A useful version of the above lemma is when applied to matrices of the
form $\Xnew\trans{\Xnew}$. We will need control over the upper and
lower singular values of such matrices for our proofs, which we next
provide.


\begin{lemma}\label{lem:Xnew-covariance-spectralnorm}
With probability at least $1-r\exp\left(-\frac{C n
}{rs}\right)-r\exp\left(-\frac{C n }{rM^2s}\right)$, for every $r
\times n$ matrix $\Xnew$ s.t. $\supp(\Xnew)\subseteq \supp(\Xstar)$,
we have:

$$\twonorm{\Xnew\trans{\Xnew} - \Xstar\trans{\Xstar}}\leq
4\left(\infnorm{\Xnew - \Xstar} + \infnorm{\Xnew -
  \Xstar}^2\right)\cdot \frac{s^2n}{r}.$$

Further assuming $\norm{\Xnew - \Xstar}_\infty \leq 1/(64s)$, we have
with the same probability

\begin{equation*}
  \singmin(\Xnew \trans{\Xnew}) \geq \frac{ns}{8r}.
\end{equation*}
\end{lemma}

\bprf

Let $\Xnew=\Xstar+E_{\Xstar}$. Note that $\supp(E_{\Xstar})\subseteq
\supp(\Xstar)$. Now,

$$\|\Xnew\trans{\Xnew}-\Xstar\trans{\Xstar}\|_2 \leq
\|E_{\Xstar}\|_2(\|E_{\Xstar}\|_2+2 \|\Xstar\|_2).$$ By
Lemma~\ref{lem:W-spectralnorm}, $\twonorm{E_{\Xstar}} \leq
2s\sqrt{\frac{n}{r}} \norm{\E_{\Xstar}}_\infty$ with probability at
least $\geq 1-r\exp\left(-\frac{C n}{rs}\right)$. Combining this with
the bound on $\twonorm{\Xstar}$ from
Lemma~\ref{lem:empiricalcovariance-concentration} completes the
proof. The second statement now follows by combining the result with
our earlier lower bound on the minimum singular value of $\Xstar$ in
Lemma~\ref{lem:empiricalcovariance-concentration}.
\eprf

A particular consequence of this lemma which will be useful is a lower
bound on the diagonal entries of the matrix $X\trans{X}$. Indeed, we
see that under the assumption $\norm{\Xnew - \Xstar}_\infty \leq
1/(64s)$, with probability at least $1-r\exp\left(-\frac{C n
}{rs}\right)-r\exp\left(-\frac{C n }{rM^2s}\right)$ we have the lower
bound uniformly for all $p = 1,2,\ldots, r$

\begin{equation}
  \row{\Xnew}{p}\trans{\row{\Xnew}{p}} \geq \frac{ns}{8r}.
  \label{eqn:element-lb-xxtrans}
\end{equation}

We finally have the following concentration lemma, which is a simple
consequence of the Bernstein concentration bound.
\begin{lemma}\label{lem:dpchernoff}
Let $\dip$ be as defined in Section~\ref{sec:altmin-mainproofs}. Then, with
probability at least $1-\exp\left(-\frac{\delta^2 ns}{3rM^2}\right)$:

\begin{enumerate}
\item $  (1-\delta) \frac{\spindex n}{r}\leq \sum_{i=1}^n \dip \leq
  (1+\delta) \frac{\spindex n}{r},\; \forall \; p \in [r]
  $, and
\item $(1-\delta) \frac{\spindex n}{r} \leq \twonorm{\col{\Xstar}{p}}^2
  = \sum_{i=1}^n \dip \left(\elt{M}{p}{i}\right)^2 \leq (1+\delta)
  \frac{\spindex n}{r}\forall \; p \in
  [r]$
\end{enumerate}
\end{lemma}

\bprf 

We start with the proof of the second part, noting that the first part
then immediately follows by setting $(\elt{M}{p}{i})^2 \equiv 1$.  The
second part will follow from a straightforward use of Bernstein's
inequality. Note that $\abs{\elt{M}{p}{i}} \leq M$ and
$\E[(\elt{M}{p}{i})^2] = 1$. As a result, for all $i = 1,2,\ldots, n$
we have $|\dip (\elt{M}{p}{i})^2| \leq M^2$, and
\begin{align*}
  \mbox{Var}[\dip (\elt{M}{p}{i})^2] \leq \E[\dip (\elt{M}{p}{i})^4]
  \leq M^2 \E[\dip (\elt{M}{p}{i})^2] = M^2.
\end{align*}

Also, we have $\E[\dip(\elt{M}{p}{i})^2] = \E[\dip] = s/r$.
Consequently, we obtain that with probability at least $1 -
\exp(-ns\delta^2/(rM^2(1 + \delta/3)))$ we have

\begin{equation*}
  \abs{\sum_{i=1}^n \dip (\elt{M}{p}{i})^2 - \frac{ns}{r}} \leq
  \frac{\delta ns}{r}.
\end{equation*}
To complete the proof, note that $1 \geq \delta/3$ which yields the
stated error probability. Finally, as stated before, we can recover
the first part by setting $(\elt{M}{p}{i})^2 \equiv 1$.  
\eprf

\begin{lemma}\label{lem:2-intersection}
  With probability at least \mbox{$1 - r\exp\left(-\frac{C
      n}{r^2}\right) - r\exp\left(-\frac{C n}{rM^2}\right)$}, for every
  $r\times n$ matrix $\Xnew$ s.t. $\supp(\Xnew)\subseteq
  \supp(\Xstar)$, we have the following bounds uniformly for all $p =
  1,2,\ldots, r$
\begin{enumerate}
  \item $\twonorm{\elt{\left(\delX \trans{\Xnew}\right)}{\setminus
      p}{p}} \leq \left(1+\sqrt{s}\infnorm{\delX}\right)
    \frac{4\sqrt{2}\infnorm{\delX} s^2 n}{r^{\frac{3}{2}}}$, and
  \item $\twonorm{\row{\Xnew}{\setminus p}
    \trans{\left(\row{\Xnew}{p}\right)}} \leq
    \left(1+\sqrt{s}\infnorm{\delX}\right)^2\frac{4s^2n}{r^{\frac{3}{2}}}$,
\end{enumerate}
where $\delX \defas \Xnew - \Xstar$.
\end{lemma}

\bprf
Since $X$ has the same sparsity pattern as $\Xstar$, we can rewrite it
as $X^p_i=\dip X^p_i$. We start by proving the first part of the
lemma.

{\bf Proof of Part 1}: Without loss of generality, we will prove the
statement for $p = 1$.  Let $D$ denote the $n\times n$ diagonal matrix
with: 
\begin{align*}
  \elt{D}{i}{i}=
	\left\{ \begin{array}{cc}
	1, &\mbox{if } \elt{\Xstar}{1}{i} \ne 0, \\
	0, &\mbox{otherwise.}
  \end{array}
\right.
\end{align*}
Using this notation, we have $\elt{\left(\delX
  \trans{\Xnew}\right)}{\setminus 1}{1} = \elt{\left(\delX D
  \trans{\Xnew}\right)}{\setminus 1}{1}$.  So, we have:

\begin{align*}
  \twonorm{\elt{\left(\delX \trans{\Xnew}\right)}{\setminus 1}{1}}
  &=   \twonorm{\elt{\left(\delX D
      \trans{\Xnew}\right)}{\setminus 1}{1}} \\
  &\leq \twonorm{\row{\left(\delX D\right)}{\setminus 1}}
  \twonorm{\col{\left(\trans{\Xnew}\right)}{1}} \\
  &\leq \twonorm{\row{\left(\delX D\right)}{\setminus 1}}
  \twonorm{\col{\left(\trans{\Xstar}\right)}{1} +
    \col{\left(\trans{\delX}\right)}{1}} \\
  &\stackrel{(\zeta_1)}{\leq} \twonorm{\row{\left(\delX
      D\right)}{\setminus 1}} \cdot \left(\sqrt{\frac{2sn}{r}} +
  \infnorm{\delX} s\sqrt{\frac{2n}{r}}\right),
\end{align*}
with probability at least $1-r\exp\left(-Cn/rs\right)$,
where the first term in $(\zeta_1)$ follows from the second part of
Lemma~\ref{lem:dpchernoff} (setting $\delta = 1$) and the second is a
consequence of Lemma~\ref{lem:W-spectralnorm}. In order to control
$\twonorm{\row{\left(\delX D\right)}{\setminus 1}}$, we observe that
it is a matrix with a random number of columns selected by the matrix
$D$. In particular, conditioned on $\set{i : \elt{D}{i}{i} = 1}$, the
support of $\elt{\Xstar}{\setminus 1}{i}$ is independent over $s-1$
sparse vectors (and the support of $\delX$ is a subset of the support
of $\Xstar$). Hence we can easily see that, 
\begin{align}
  \P\left[ \twonorm{\row{\left(\delX D\right)}{\setminus 1}} > t
    \right] &\leq \P\left[ \twonorm{\row{\left(\delX
        D\right)}{\setminus 1}} > t \cap \frac{\spindex n}{2r} <
    \abs{\set{i : \elt{D}{i}{i} = 1}} < \frac{2\spindex n}{r} \right]
  \nonumber\\
  &\qquad \qquad + \P\left[ \abs{\set{i : \elt{D}{i}{i} = 1}} \leq
    \frac{\spindex n}{2r} \cup  \abs{\set{i : \elt{D}{i}{i} = 1}} \geq
    \frac{2\spindex n}{r} \right].
  \label{eqn:delXD-prob}
\end{align}
In order to control the first probability, we note that
\begin{align*}
  &\P\left[ \twonorm{\row{\left(\delX
        D\right)}{\setminus 1}} > t \cap \frac{\spindex n}{2r} <
    \abs{\set{i : \elt{D}{i}{i} = 1}} < \frac{2\spindex n}{r}
    \right]\\
  &= \sum_{m = \lfloor sn/2r \rfloor}^{\lceil 2sn/r \rceil} \P\left[
    \twonorm{\row{\left(\delX D\right)}{\setminus 1}} > t \cap
    \abs{\set{i : \elt{D}{i}{i} = 1}} = m\right]\\
  &= \sum_{m = \lfloor sn/2r \rfloor}^{\lceil 2sn/r \rceil} \P\left[
    \twonorm{\row{\left(\delX D\right)}{\setminus 1}} > t \mid
    \abs{\set{i : \elt{D}{i}{i} = 1}} = m\right] \P\left[ \abs{\set{i
        : \elt{D}{i}{i} = 1}} = m\right].
\end{align*}
Setting $t = 2\infnorm{\delX}\sqrt{s^2m/r}$, we obtain as a
consequence of Lemma~\ref{lem:W-spectralnorm}: 
\begin{align*}
  &\P\left[ \twonorm{\row{\left(\delX
        D\right)}{\setminus 1}} > t \cap \frac{\spindex n}{2r} <
    \abs{\set{i : \elt{D}{i}{i} = 1}} < \frac{2\spindex n}{r}
    \right]\\
  &= \sum_{m = \lfloor sn/2r \rfloor}^{\lceil 2sn/r \rceil}
  r\exp\left(-\frac{Cm}{rs} \right) \P\left[ \abs{\set{i
        : \elt{D}{i}{i} = 1}} = m\right]\\
  &\leq r\exp\left(-\frac{Cn}{2r^2} \right).
\end{align*}
The second probability in Equation~\ref{eqn:delXD-prob} can be bounded
through part 1 of Lemma~\ref{lem:dpchernoff}, since

\begin{equation*}
  \abs{\set{i : \elt{D}{i}{i} = 1}} = \sum_{i=1}^n
  \elt{\spmat}{1}{i}.
\end{equation*}
Doing so, we obtain with probability at least \mbox{$1 -
  r\exp\left(-\frac{C n}{2r^2}\right) -
  r\exp\left(-Cns/(3rM^2)\right)$}: 
\begin{align*}
  \twonorm{\row{\left(\delX D\right)}{\setminus 1}} \leq
  2\infnorm{\delX} \spindex \sqrt{\frac{\left(\frac{2\spindex
        n}{r}\right)}{r}}.
\end{align*}
This proves part 1.

\noindent {\bf Proof of Part 2}: The proof of this is similar to that of part
1. Wlog, assume $p=1$. We have:

\begin{align*}
  \twonorm{\row{\Xnew}{\setminus 1} \trans{\left(\row{\Xnew}{1}\right)}}
  &=  \twonorm{\row{\Xnew}{\setminus 1} D
    \trans{\left(\row{\Xnew}{1}\right)}} \\
  &\leq \twonorm{\row{\left(\Xnew D\right)}{\setminus 1}}
  \twonorm{\col{\left(\trans{\Xnew}\right)}{1}} \\
  &\stackrel{}{\leq} \twonorm{\row{\left(\Xnew D\right)}{\setminus 1}}
  \cdot 2\left(1+\sqrt{s}\infnorm{\delX}\right)\sqrt{\frac{sn}{r}}.
\end{align*}

For the first term above, we have:
\begin{align*}
  \twonorm{\row{\left(\Xnew D\right)}{\setminus 1}}
&\leq \twonorm{\row{\left(\Xstar D\right)}{\setminus 1}} +
  \twonorm{\row{\left(\delX D\right)}{\setminus 1}}.
\end{align*}

The second term in this decomposition was controlled above and the
first one can be similarly bounded. Doing so, we obtain with
probability at least \mbox{$1 - r\exp\left(-\frac{C n}{r^2}\right) -
  r\exp\left(-\frac{C n }{rM^2s}\right)$}:
\begin{align*}
\twonorm{\row{\left(\Xnew D\right)}{\setminus 1}} \leq 2\spindex
\left(1+\sqrt{s}\infnorm{\delX}\right) \sqrt{\frac{2\spindex n}{r^3}}.
\end{align*}
This proves the lemma.
\eprf

We begin with an auxiliary result on the RIP constant of an incoherent matrix.

\begin{lemma}\label{lem:RIP-A}
  Suppose $\Astar$ satisfies Assumption~$(B1)$.
  Then, the $2s$-RIP constant of $\Astar$, $\delta_{2\spindex}$ satisfies
  $\delta_{2\spindex} < \frac{2\mu_0 \spindex}{\sqrt{d}}$.
\end{lemma}
\bprf
Consider a $2\spindex$-sparse unit vector $w \in \R^r$ with $\textrm{Supp}(w)=S$. We have:
\begin{align*}
  \norm{Aw}^2 = \left(\sum_{j \in S} w_j \col{\Astar}{j}\right)^2
	&= \sum_j w_j^2 \norm{\col{\Astar}{j}}^2 + \sum_{j,l\in S,j
    \neq l} w_j w_l \iprod{\col{\Astar}{j}}{\col{\Astar}{l}} \\
	&\geq 1 - \sum_{j,l\in S,j \neq l} \abs{w_j w_l}
  \abs{\iprod{\col{\Astar}{j}}{\col{\Astar}{l}}} \\
	&\geq 1 - \sum_{j,l\in S,j \neq l} \abs{w_j w_l} \frac{\mu_0}{\sqrt{d}} \\
	&\geq 1 - \frac{\mu_0}{\sqrt{d}} \onenorm{w}^2 \\
	&\geq 1 - \frac{\mu_0}{\sqrt{d}} 2s \cdot \norm{w}_2^2 = 1 - \frac{2\mu_0 s}{\sqrt{d}}.
\end{align*}
Similarly, we have:
\begin{align*}
  \norm{\Astar w}_2^2 \leq 1 + \frac{2\mu_0 s}{\sqrt{d}}.
\end{align*}
This proves the lemma.
\eprf

\begin{lemma}\label{lem:Schurcomplement}
  We have the following formula for matrix inversion:
\begin{align*}
  \inv{\left[ \begin{array}{cc} A & B \\
        C & D\end{array}\right]}
  = \left[ \begin{array}{cc}
      \inv{A} + \inv{A}BMC\inv{A} & - \inv{A}BM \\
      -MC\inv{A} & M \end{array} \right],
\end{align*}
where $\inv{M} \defas \left(D - C \inv{A} B\right)$ is the Schur
complement of $A$ in the above matrix.
\end{lemma}


\end{document}